\documentclass{article}

\usepackage{microtype}
\usepackage{graphicx}
\usepackage{booktabs} 

\usepackage{subcaption}
\usepackage{multirow}
\usepackage{xcolor}
\usepackage{pifont}
\newcommand{\cmark}{\ding{51}}%
\newcommand{\xmark}{\ding{55}}%

\newcommand{\Comment}[1]{{\hskip2em$\blacktriangleright$ #1}}

\usepackage{hyperref}

\usepackage[accepted]{icml2023}

\usepackage{amsmath}
\usepackage{amssymb}
\usepackage{mathtools}
\usepackage{amsthm}

\usepackage[capitalize,noabbrev]{cleveref}

\theoremstyle{plain}

\theoremstyle{definition}

\theoremstyle{remark}

\icmltitlerunning{BiRT: Bio-inspired Replay in Vision Transformers for Continual Learning}

\begin{document}

\twocolumn[
\icmltitle{BiRT: Bio-inspired Replay in Vision Transformers for Continual Learning}



\icmlsetsymbol{equal}{*}

\begin{icmlauthorlist}
\icmlauthor{Kishaan Jeeveswaran}{nie}
\icmlauthor{Prashant Bhat}{nie,tue}
\icmlauthor{Bahram Zonooz}{equal,nie,tue}
\icmlauthor{Elahe Arani}{equal,nie,tue}
\end{icmlauthorlist}

\icmlaffiliation{tue}{Dep. of Mathematics and Computer Science, Eindhoven University of Technology, Netherlands}
\icmlaffiliation{nie}{Advanced Research Lab, NavInfo Europe, Netherlands}

\icmlcorrespondingauthor{}{kishaan96@gmail.com}
\icmlcorrespondingauthor{}{p.s.bhat@tue.nl, b.zonooz@tue.nl, e.arani@tue.nl}

\icmlkeywords{catastrophic forgetting, vision transformers, representation rehearsal, continual learning, brain-inspired learning mechanism, noise, dual memory}


\vskip 0.3in
]



\printAffiliationsAndNotice{\icmlEqualContribution} 

\begin{abstract}
The ability of deep neural networks to continually learn and adapt to a sequence of tasks has remained challenging due to catastrophic forgetting of previously learned tasks. Humans, on the other hand, have a remarkable ability to acquire, assimilate, and transfer knowledge across tasks throughout their lifetime without catastrophic forgetting. The versatility of the brain can be attributed to the rehearsal of abstract experiences through a complementary learning system. However, representation rehearsal in vision transformers lacks diversity, resulting in overfitting and consequently, performance drops significantly compared to raw image rehearsal. Therefore, we propose BiRT, a novel representation rehearsal-based continual learning approach using vision transformers. Specifically, we introduce constructive noises at various stages of the vision transformer and enforce consistency in predictions with respect to an exponential moving average of the working model. Our method provides consistent performance gain over raw image and vanilla representation rehearsal on several challenging CL benchmarks, while being memory efficient and robust to natural and adversarial corruptions.
\footnote{Code available at \href{https://github.com/NeurAI-Lab/BiRT}{github.com/NeurAI-Lab/BiRT}.}
\end{abstract}

\section{Introduction}
\label{sec:intro}

\begin{figure}[t]
\centering
\vskip 0.2in
  \centerline{\includegraphics[width=\linewidth]{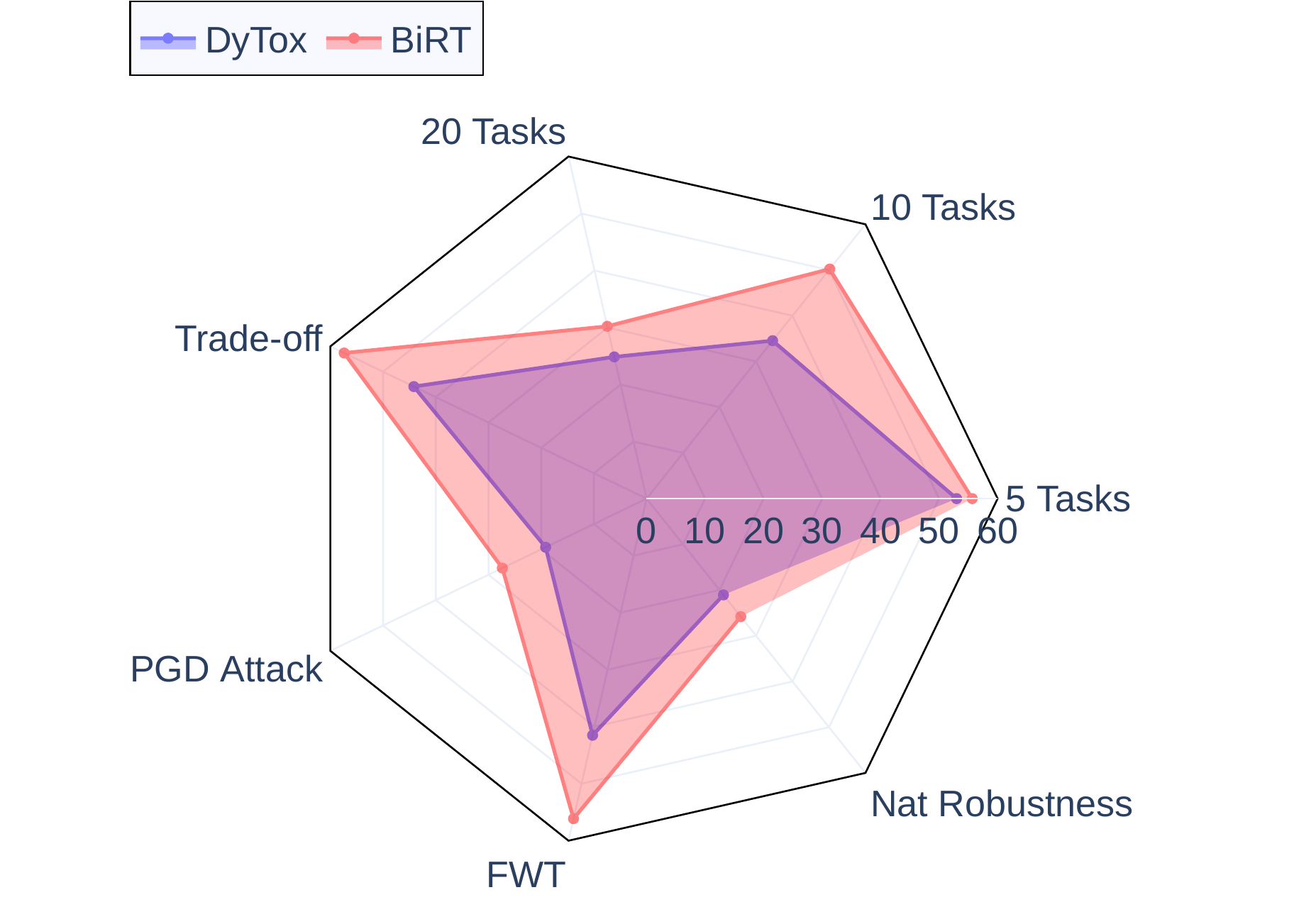}}
  \caption{Overall performance of our proposed method, BiRT, vs. DyTox trained continually on CIFAR-100 with 500 buffer size on different metrics; Top-1 accuracy is reported for all metrics. Therefore, a CL method
with full coverage of the octagon has all the ideal features: highest accuracy (on varying task sequences), natural/adversarial robustness, forward transfer, and stability-plasticity trade-off.}
  \label{fig:memroy_efficiency}
\vskip -0.2in
\end{figure}

Computational systems operating in the real world are normally exposed to a sequence of multiple tasks with non-stationary data streams.  
Similar to biological organisms, it is desirable for these artificial systems to be able to learn on a continual basis to successfully act and adapt to new scenarios in the real world. 
However, deep neural networks (DNNs) are inherently designed for training on stationary, independent, and identically distributed (i.i.d.) data. 
The sequential nature of continual learning (CL) violates this strong assumption, leading to catastrophic forgetting of older tasks. 
Catastrophic forgetting often leads to a rapid decline in the performance of old tasks and, in the worst case, the previously acquired information is completely overwritten by the new one~\citep{parisi2019continual}.

\begin{figure*}[t]
\vskip 0.2in
\begin{center}
  \centerline{\includegraphics[trim=0 0 1.5cm 0, clip, width=.98\linewidth]{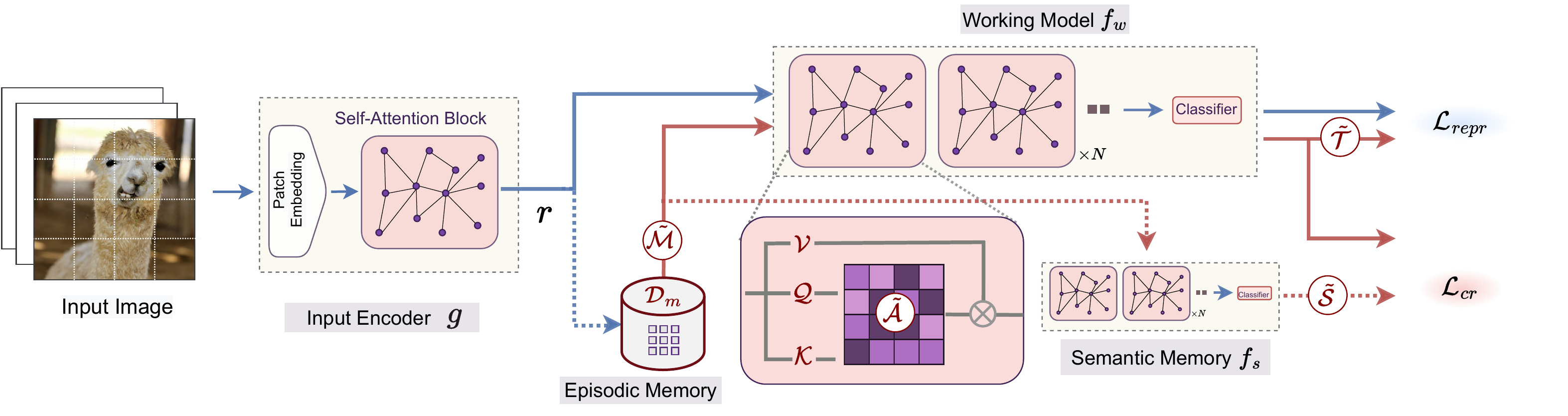}}
  \caption{\textit{BiRT} employs a bio-inspired non-veridical experience replay in a dual memory system based on vision transformers. The semantic memory, $f_s$, gradually assimilates learned knowledge from working model, $f_w$, by taking an exponential moving average over its weights. The semantic memory interacts with the episodic memory which stores the learned representations of the previous tasks ($r$). To effectively replay these abstract high-level representations, we inject constructive noise by mixing up representations ($\tilde{\mathcal{M}}$), adding noise to the internal attention maps ($\tilde{\mathcal{A}}$), and emulating trial-to-trial variability through adding noise to the outputs of semantic memory ($\tilde{\mathcal{S}}$) and to the targets ($\tilde{\mathcal{T}}$). To retrieve the knowledge, the consolidated knowledge from semantic memory is enforced to the working model in the functional space via a consistency regularization.}
  \label{fig:representation_replay}
\end{center}
\vskip -0.2in
\end{figure*}

Rehearsal-based approaches, which store and replay previous task samples, have been fairly successful in mitigating catastrophic forgetting in CL. Recent evidence suggests that replay might even be unavoidable in certain CL scenarios \cite{farquhar2018towards}. However, replaying raw pixels from past experiences is not consistent with neurophysiological mechanisms in the brain \citep{kudithipudi2022biological,hayes2019memory}. Furthermore, the replay of raw pixels is memory inefficient and raises data privacy and security concerns \citep{mai2022online}.
Juxtaposing biological and artificial experience rehearsal, representation rehearsal is a lucrative alternative to address the problems associated with raw image rehearsal in CL.  Representation rehearsal, either generative \cite{van2020brain, lao2020continuous} or by storing \cite{hayes2020remind, caccia2020online, iscen2020memory},  entails replaying the latent features of the intermediate layers of DNNs to mitigate catastrophic forgetting. In generative methods, the generator itself is as large as the CL model and is prone to catastrophic forgetting. Additionally, generative models are difficult to train and suffer mode collapse. However, although storing representations is memory and computation efficient, choosing an ideal layer for rehearsal remains an open question. Furthermore, stored representations in a bounded memory lack diversity, resulting in overfitting.

In contrast, the human brain learns, stores, and remembers experiences without catastrophically forgetting previous tasks. The versatility of the brain can be attributed to the rehearsal of abstract experiences through multiple memory systems \citep{hassabis2017neuroscience} and a rich set of neurophysiological processing principles \citep{parisi2019continual}. In addition, the brain harbors random disturbances of signals, termed noise, that contribute to cellular and behavioral trial-to-trial variability \cite{faisal2008noise}. Although noise is sometimes considered a nuisance, noise forms a notable component of the computational strategy of the brain. The brain exploits noise to perform tasks, such as probabilistic inference through sampling, that facilitate learning and adaptation in dynamic environments \cite{maass2014noise}. As is the case in the brain, we hypothesize that noise can be a valuable tool in improving generalization in representation rehearsal in vision transformers.

To this end, we propose BiRT, a novel representation rehearsal-based continual learning method based on vision transformers, architectures composed of self-attention modules inspired by human visual attention \citep{lindsay2020attention}.  Specifically, our method consists of two complementary learning systems: a working model and semantic memory, an exponential moving average of the working model. To reduce overfitting and bring diversity in representation rehearsal, BiRT introduces various controllable noises at various stages of the vision transformer and enforces consistency in predictions with respect to semantic memory. As semantic memory consolidates semantic information, consistency regularization in the presence of meaningful noise promotes generalization while effectively reducing overfitting. BiRT provides a consistent performance gain over the raw image and the vanilla representation rehearsal on several CL scenarios and metrics while being robust to natural and adversarial corruptions (Figure \ref{fig:memroy_efficiency}).

\section{Related Work}

\textbf{Continual Learning}: 
DNNs are typically designed to incrementally adapt to stationary i.i.d. data streams shown in isolation and random order \citep{parisi2019continual}. Therefore, sequential learning over non-i.i.d. data causes catastrophic forgetting of previous tasks and overfitting of the current task. 
Approaches to address catastrophic forgetting can be broadly divided into three categories: regularization-based approaches \citep{kirkpatrick2017overcoming, zenke2017continual, li2017learning} penalize changes in important parameters pertaining to previous tasks, parameter isolation methods \citep{rusu2016progressive, aljundi2017expert, fernando2017pathnet} allocate a distinct set of parameters for distinct tasks, and rehearsal-based approaches \citep{ratcliff1990connectionist, rebuffi2017icarl, lopez2017gradient, bhat2023taskaware} store old task samples and replay them alongside current task samples. Among different approaches to mitigate catastrophic forgetting, 
experience rehearsal is fairly successful in multiple CL scenarios \citep{parisi2019continual}.

Rehearsal-based approaches replay raw pixels from past experiences, inconsistent with how humans continually learn \citep{kudithipudi2022biological}. Furthermore, the replay of raw pixels can have other ramifications, including a large memory footprint, data privacy, and security concerns \citep{mai2022online}.
Therefore, several works \citep{pellegrini2020latent, iscen2020memory, caccia2020online} mimic abstract representation rehearsal in the brain by storing and replaying representations from intermediate layers in DNNs. Representation rehearsal can be done by employing generative models \cite{van2020brain, lao2020continuous}  or by storing previous task representations in the buffer \cite{hayes2020remind, iscen2020memory}. While generative models themselves are prone to forgetting and mode collapse, storing representations in a bounded memory buffer lacks diversity due to the unavailability of proper augmentation mechanisms. Although high-level representation replay can potentially mitigate memory overhead and privacy concerns, replaying representations over and over again leads to overfitting.

\textbf{Transformers for CL:} Transformer architectures \citep{vaswani2017attention} were first developed for machine translation and later expanded to computer vision tasks \citep{dosovitskiy2020image, touvron2021training, jeeveswaran2022comprehensive} by considering image patches as replacements for tokens. Despite their success in several benchmarks, vision transformers have not been widely considered for continual learning. \citet{yu2021improving} studied transformers in a class-incremental learning setting and pointed out several problems in naively applying transformers in CL. DyTox \citep{douillard2021dytox} proposed a dynamically expanding architecture using separate task tokens to model the context of different classes in CL. LVT \citep{wang2022continual} proposed an external key and an attention bias to stabilize the attention map between tasks and used a dual classifier structure to avoid catastrophic interference while learning new tasks. \citet{pelosin2022towards} proposed an asymmetric regularization loss on pooled attention maps with respect to the model learned on the previous task to continually learn in an exemplar-free approach. Several other concurrent works \citep{ermis2022continual, wang2022learning, wang2022dualprompt} harnessed the pre-trained model and incorporated the learning of generic and task-specific parameters. Unlike these works, we do not use pre-trained models and replay intermediate representations instead of raw image inputs.

We seek to improve the performance of vision transformers under representation rehearsal in CL. As noise plays a constructive role in the brain, we mimic the prevalence of noise in the brain and the consequent trial-to-trial variability by injecting noise into our proposed method.

\section{Proposed Method}

The CL paradigm normally consists of $T$ sequential tasks, with the data gradually becoming available over time. During each task $t \in \{1, 2, .., T\}$, the samples and the corresponding labels $(x_i, y_i)_{i=1}^{N}$ are drawn from the task-specific distribution $\mathcal{D}_t$. The continual learning model $f_\theta$ is optimized sequentially on one task at a time, and inference is carried out on all the tasks seen so far. CL is especially challenging for vision transformers due to the limited training data for every task \citep{raghu2021vision, touvron2021training} in addition to the issue of catastrophic forgetting. By mimicking the association of past and present experiences in the brain, experience rehearsal (ER) partially addresses the problem of catastrophic forgetting. Thus, the learning objective of ER is as follows:
\begin{align}
\begin{split}
\label{eqn_er}
 \mathcal{L}_{er} \triangleq  \displaystyle &\mathop{\mathbb{E}}_{(x_{i}, y_{i}) \sim \mathcal{D}_{t}} \left[ \: \mathcal{L}_{ce} (f_\theta({x}_i), y_i) \: \right] \\
 &+ \alpha \displaystyle \mathop{\mathbb{E}}_{(x_j, y_j) \sim \mathcal{D}_{m}} \left[ \; \mathcal{L}_{ce} (f_\theta({x}_j), y_j) \; \right],
\end{split}
\end{align}
where $\alpha$ represents a balancing parameter, $\mathcal{D}_{m}$ is episodic memory, and $\mathcal{L}_{ce}$ is cross-entropy loss. 
To further reduce catastrophic forgetting, we employ a complementary learning system based on abstract, high-level representation rehearsal. 
To promote diversity and generalization in representation rehearsal, we introduce various controllable noises at different stages of the vision transformer and enforce consistency in predictions with respect to the semantic memory. In the following sections, we describe in detail different components of BiRT.

\subsection{Knowledge Consolidation through complementary learning system}
Complementary learning system (CLS) theory posits that the hippocampus and neocortex entail complementary properties necessary to capture complex interactions in the brain \citep{mcnaughton1995there}. Inspired by CLS theory, we propose a dual memory transformer-based learning system that acquires and assimilates knowledge over short and long periods of time. The working model encounters new tasks and consolidates knowledge over short periods of time. We then gradually aggregate the weights of the working model into semantic memory during intermittent stages of inactivity. Following \citet{arani2022learning}, we design the semantic memory as an exponential moving average of the working model as follows:
\begin{equation}
\label{eqn_cls}
 \theta_{s} = \gamma \theta_{s} + (1 - \gamma) \theta_{w}
\end{equation}
where $\theta_{w}$ and $\theta_{s}$ are the weights of the working model and semantic memory, respectively, and $\gamma$ is a decay parameter. As the working model focuses on specializing on the current task,  the copy of the working model at each training step can be considered as an expert on a particular task. Therefore, the aggregation of weights throughout CL training can be deemed as an ensemble of expert models that consolidate knowledge across tasks, resulting in smoother decision boundaries.

\subsection{Episodic Memory}
In line with experience rehearsal in the brain \citep{ji2007coordinated}, we propose an abstract, high-level representation rehearsal for vision transformers. The working model comprises two nested functions: $g(.)$ and $f_{w}(.)$. The first few layers of the encoder, $g(.)$, process the raw image input, and the output along with the ground truth label is stored in episodic memory $\mathcal{D}_m$.
To ensure consistency in intermediate representations, $g(.)$ can be initialized using pre-trained weights and fixed before starting CL training or fixed after learning some tasks.
On the other hand, $f_{w}(.)$, the later layers of the transformer, process abstract high-level representations, and remain learnable throughout the CL training. During intermittent stages of inactivity, the stable counterpart semantic memory $f_{s}(.)$ is updated according to Eq. \ref{eqn_cls}.

The episodic memory is populated at the task boundary using iCaRL herding \citep{rebuffi2017icarl}. Representations $r_j = g(x_j)$, stored in episodic memory, are interleaved with current task representations and are processed synchronously by $f_{w}(.)$ and $f_{s}(.)$. The learning objective for representation rehearsal can thus be obtained by adapting Eq. \ref{eqn_er} as follows:
\begin{align}
\begin{split}
\label{eqn_repr}
 \mathcal{L}_{repr} \triangleq \displaystyle \mathop{\mathbb{E}}_{(x_{i}, y_{i}) \sim \mathcal{D}_{t}} \left[ \: \mathcal{L}_{ce} (f_w(g(x_i)), y_i) \: \right] \\
 + \; \alpha \displaystyle \mathop{\mathbb{E}}_{(r_j, y_j) \sim \mathcal{D}_{m}} \left[ \; \mathcal{L}_{ce} (f_w(r_j), y_j) \; \right]
\end{split}
\end{align}

\begin{algorithm}[tb]
\caption{BiRT Algorithm}
\label{alg:method}
\begin{algorithmic}
\STATE \textbf{input:} Data streams $\mathcal{D}_{t}$, buffer $\mathcal{D}_{m}$, working model $f_w$, hyperparameters $\gamma$, $\alpha_{t}$, $\alpha_{m}$, $\alpha_{a}$, $\alpha_{s}$
\FORALL {tasks $t \in \{1, 2,..,T\}$} 
    \FOR {epochs $e \in \{1, 2,..,E\}$}
      \STATE sample a mini-batch ${(x, y)} \sim \mathcal{D}_{t}$
      \STATE $x = augment(x)$ 
      \IF{$\mathcal{D}_{m} \neq \emptyset$}
         \STATE sample a mini-batch ${(r, y)} \sim \mathcal{D}_{m}$ 
         \STATE{$a, b, c, d, e \sim \mathcal{U}(0, 1)$}
         \STATE $ \tilde{y} \gets \tilde{\mathcal{T}}(y)$ ~~~~~~~~~~~~~~~~~\textbf{if} $a < \alpha_{t}$
         \STATE $(\tilde{r}, \tilde{y}) \gets \tilde{\mathcal{M}}(r,y)$ ~~~~~\textbf{if} $b < \alpha_{m}$\hfill \Comment{\small (Eq. \ref{eqn_mixup})}
         \STATE $ \tilde{A} \gets \tilde{\mathcal{A}}(A)$ ~~~~~~~~~~~~~~~\textbf{if} $c < \alpha_{a}$\hfill \Comment{\small (Eq. \ref{eq:attention_noise})}
         \STATE $f_s(r) \gets \tilde{\mathcal{S}}(f_s(r), \delta)$ ~\textbf{if} $d < \alpha_{s}$\hfill \Comment{\small (Eq. \ref{eqn_logitn})}
     \ENDIF
 
      \STATE Compute outputs of $f_w(.)$ and $f_s(.)$ 
      \STATE Compute $\mathcal{L} = \mathcal{L}_{repr} + \rho \mathcal{L}_{cr} \;$ \hfill \Comment{\small (Eqs. \ref{eqn_repr}, \ref{eqn_cr}, \ref{eqn_final})}
      \STATE $\theta_{w} \gets \theta_{w} + \nabla_{\theta_{w}} \mathcal{L}$ 
      \STATE $\theta_{s} \gets \gamma \theta_{s} + (1 - \gamma) \theta_{w}$ ~\textbf{if} $e < \alpha_{e}$ \textbf{and} $t > 1$
    \ENDFOR
    \IF{task-end = True}
      \IF{t = 1}
         \STATE Freeze $g(.)$
         \STATE $\theta_{s} =  \text{copy}(\theta_{w})$
      \ENDIF
      \STATE $\mathcal{D}_{m} \gets (r, y)$
    \ENDIF
\ENDFOR
\STATE \textbf{Return:}{working model $\theta_w$, and semantic memory $\theta_s$}
\end{algorithmic}
\end{algorithm}

\subsection{Noise and Trial-to-Trial Variability}

Noise is prevalent at every level of the nervous system and has recently been shown to play a constructive role in the brain \citep{faisal2008noise, mcdonnell2011benefits}. Trial-to-trial variability, a common phenomenon in biological systems in which the neural response to the same stimuli differs across trials, is often the result of noise \citep{faisal2008noise}.
Trial-to-trial variability has been shown to be one of the key components of the computational mechanism in the brain \citep{maass2014noise}.
Furthermore, injecting noise into the neural network learning pipeline has been shown to result in faster convergence to the global optimum \citep{zhou2019toward}, better generalization \citep{srivastava2014dropout}, and effective knowledge distillation.

To simulate noise and trial-to-trial variability, we stochastically inject constructive noise into various components of our CL setup. In the following sections, we describe in detail how exactly we leverage noise during CL training.

\begin{table*}[t]
\centering
\caption{Results on multiple datasets learned with 10 tasks with varying buffer sizes, averaged over multiple class orders. BiRT achieves consistent improvements over DyTox in different metrics, i.e. accuracy, forgetting, BWT, and FWT. The last accuracy determines the performance on past tasks after learning the last task, and the average accuracy shows the average of the last accuracy after learning every task.}
\label{tab:main}
\vskip 0.05in 
\begin{small}
\begin{sc}
\resizebox{\textwidth}{!}{
\begin{tabular}{@{}llccccccc@{}}
\toprule
& \multicolumn{2}{l}{Buffer Size}& \multicolumn{2}{c}{500} & \multicolumn{2}{c}{1000} & \multicolumn{2}{c}{2000} \\
& & Joint& DyTox & BiRT & DyTox & BiRT & DyTox & BiRT \\
\midrule

\multirow{5}{*}{\textbf{CIFAR-100}}
& Last Acc $\uparrow$     & 74.99\tiny{$\pm$0.22} & 34.54\tiny{$\pm$1.82}& \textbf{50.20}\tiny{$\pm$0.67}& 43.92\tiny{$\pm$0.84} & \textbf{51.20}\tiny{$\pm$1.46} & 52.34\tiny{$\pm$0.46} & \textbf{53.01}\tiny{$\pm$0.57}\\
& Avg Acc  $\uparrow$     & & 58.35\tiny{$\pm$1.54 } & \textbf{63.82}\tiny{$\pm$1.80}& 63.67\tiny{$\pm$1.31} & \textbf{64.56}\tiny{$\pm$2.31} & \textbf{68.42}\tiny{$\pm$1.13} & 66.70\tiny{$\pm$0.36} \\
& BWT $\uparrow$          & & -39.79\tiny{$\pm$1.16} & \textbf{-15.62}\tiny{$\pm$0.29} & -32.05\tiny{$\pm$0.33} & \textbf{-15.25}\tiny{$\pm$0.66} & -24.44\tiny{$\pm$0.65} & \textbf{-16.30}\tiny{$\pm$1.31} \\
& FWT $\uparrow$          & & 41.51\tiny{$\pm$1.61}& \textbf{56.14}\tiny{$\pm$1.52}& 50.04\tiny{$\pm$1.17} & \textbf{57.04}\tiny{$\pm$2.2} & 57.77\tiny{$\pm$0.77} & \textbf{59.74}\tiny{$\pm$1.30}\\
& Forgetting $\downarrow$ & & 53.87\tiny{$\pm$1.95}& \textbf{17.45}\tiny{$\pm$0.61}& 43.64\tiny{$\pm$0.71} & \textbf{17.70}\tiny{$\pm$1.42} & 33.92\tiny{$\pm$0.79} & \textbf{19.00}\tiny{$\pm$1.98}  \\
\midrule
\multirow{5}{*}{\textbf{TinyImageNet}}  
& Last Acc $\uparrow$& 58.46\tiny{$\pm$0.60} &23.95\tiny{$\pm$0.71} & \textbf{32.60}\tiny{$\pm$.018} & 33.25\tiny{$\pm$1.28} & \textbf{38.41}\tiny{$\pm$0.33} & 37.34\tiny{$\pm$0.22} & \textbf{40.49}\tiny{$\pm$0.52} \\
& Avg Acc $\uparrow$ &                       &42.53\tiny{$\pm$1.74} & \textbf{44.57}\tiny{$\pm$2.84}& 48.74\tiny{$\pm$1.29} & \textbf{49.26}\tiny{$\pm$2.34} & \textbf{51.30}\tiny{$\pm$2.17} & 51.15\tiny{$\pm$0.34}\\
& BWT $\uparrow$ &                           & -40.46\tiny{$\pm$0.41}&\textbf{ -13.38}\tiny{$\pm$0.98} & -31.12\tiny{$\pm$1.19} & \textbf{-17.34}\tiny{$\pm$0.51 } & -27.68\tiny{$\pm$0.77} & \textbf{-17.85}\tiny{$\pm$0.37}\\
& FWT $\uparrow$ &                           &27.84\tiny{$\pm$1.02} & \textbf{37.87}\tiny{$\pm$1.91} & 36.60\tiny{$\pm$0.34} & \textbf{41.97}\tiny{$\pm$1.54} & 40.39\tiny{$\pm$1.16} & \textbf{43.93}\tiny{$\pm$1.54}\\
& Forgetting $\downarrow$ &                  &52.32\tiny{$\pm$0.94} & \textbf{14.57}\tiny{$\pm$2.00} & 40.07\tiny{$\pm$2.12} & \textbf{18.85}\tiny{$\pm$0.22}  & 35.56\tiny{$\pm$1.29} & \textbf{19.48}\tiny{$\pm$0.21}\\
\midrule
\multirow{5}{*}{\textbf{ImageNet-100}} 
& Last Acc $\uparrow$& 79.05\tiny{$\pm$0.16} &  39.03\tiny{$\pm$1.57} & \textbf{51.05}\tiny{$\pm$0.24} & 50.62\tiny{$\pm$1.04} & \textbf{52.89}\tiny{$\pm$0.96} & 58.54\tiny{$\pm$0.42} & \textbf{59.52}\tiny{$\pm$1.39}\\
& Avg Acc $\uparrow$ & &  60.52\tiny{$\pm$1.56} & \textbf{65.51}\tiny{$\pm$0.30}& 68.14\tiny{$\pm$1.38} & \textbf{67.33}\tiny{$\pm$0.57}  & \textbf{71.67}\tiny{$\pm$1.71} & 70.51\tiny{$\pm$1.87}\\
& BWT $\uparrow$ & & -38.15\tiny{$\pm$0.48}& \textbf{-14.42}\tiny{$\pm$0.06} & -26.87\tiny{$\pm$0.72} & \textbf{-12.90}\tiny{$\pm$0.31} & -21.10\tiny{$\pm$0.78} & \textbf{-16.53}\tiny{$\pm$0.84} \\
& FWT $\uparrow$ & &  44.94\tiny{$\pm$1.69} & \textbf{58.27}\tiny{$\pm$0.30}& 56.86\tiny{$\pm$1.46} & \textbf{60.78}\tiny{$\pm$0.86}& 62.85\tiny{$\pm$1.54} & \textbf{63.40}\tiny{$\pm$2.01} \\
& Forgetting $\downarrow$ & &  51.71\tiny{$\pm$0.91} & \textbf{16.10}\tiny{$\pm$0.42}& 37.93\tiny{$\pm$11.23} & \textbf{14.83}\tiny{$\pm$0.67} & 28.68\tiny{$\pm$1.41} & \textbf{19.79}\tiny{$\pm$0.61} \\
\bottomrule
\end{tabular}}
\end{sc}
\end{small}
\vskip -0.1in
\end{table*}

\subsubsection{Representation noise $\mathcal{\Tilde{M}}$}

During CL training, the working model encounters task-specific data $\mathcal{D}_t$ that are first fed into $g(.)$, and then the output representations of $g(.)$ are interleaved with the representations of previous task samples from episodic memory $\mathcal{D}_m$. We update $\mathcal{D}_m$ at the task boundary using iCaRL herding. The interleaved representations are then processed by both $f_{w}(.)$ and $f_{s}(.)$.
Analogous to the replay of novel samples in the brain \citep{liu2019human}, we linearly combine representations sampled from episodic memory using a manifold mixup \citep{verma2019manifold}:
\begin{align}
\begin{split}
\label{eqn_mixup}
 \tilde{r} &= \lambda  r_i + (1-\lambda) \; r_j \\
 \tilde{y} &= \lambda  y_i + (1-\lambda) \;  y_j,
\end{split}
\end{align}
where $r_i, r_j$ are stored representations of two different samples and $y_i, y_j $ are the corresponding labels. Here, the mixing coefficient $\lambda $ is drawn from a Beta distribution.  As manifold mixup interpolates representations of samples belonging to different classes / tasks, it brings diversity for the experience-rehearsal, thereby reducing overfitting. 

\subsubsection{Attention noise $\mathcal{\Tilde{A}}$}
As we employ vision transformer as our architecture of choice, self-attention forms the core component of BiRT. The working model $f_{w}(.)$ in BiRT consists of several multi-head self-attention layers that map a query and a set of key-value pairs to an output. We inject noise into the scaled dot-product attention at each layer of $f_{w}(.)$ while replaying the representation as follows:
\begin{equation}
\label{eq:attention_noise}
\operatorname{Attention}(Q, K, V)=(\operatorname{softmax}\left(\frac{Q K^T}{\sqrt{d_k}}\right) + \epsilon ) \; V
\end{equation}
where $Q$, $K$ and $V$ are query, key and value matrices, and $\epsilon \sim \mathcal{N}(0, \sigma^2)$ is a white Gaussian noise. By stochastically injecting noise into self-attention, we discourage BiRT from attending to sample specific features, thereby potentially mitigating overfitting.  

\subsubsection{Supervision noise $\mathcal{\Tilde{T}}$ and $\mathcal{\Tilde{S}}$ }
We now shift our focus toward the supervision signals to further reduce overfitting in CL. Due to over-parameterization, the CL model tends to overfit on the limited number of samples from the buffer. Therefore, we introduce a synthetic label noise ($\mathcal{\Tilde{T}}$) wherein a small percentage of the samples are re-assigned a random class. BiRT takes advantage of the fact that label noise is sparse, meaning that only a fraction of the labels are corrupted while the rest are intact in the real world \cite{liu2022robust}. In addition, the harmful effects of inherent label noise on generalization can be mitigated by using additional controllable label noise \cite{chen2021noise}.

During intermittent stages of inactivity, the knowledge in the working model is consolidated into semantic memory through Eq. \ref{eqn_cls}. Therefore, knowledge of previous tasks is encoded in semantic memory weights during the learning trajectory of the working model \citep{hinton2015KD}. Then, to retrieve the structural knowledge encoded in the semantic memory, we regularize the function learned by the working model by enforcing consistency in its predictions with respect to the semantic memory:
\begin{align}
\begin{split}
\label{eqn_cr}
 \mathcal{L}_{cr} \triangleq \beta_1 &\displaystyle \mathop{\mathbb{E}}_{x_i \sim D_{t} } \lVert f_w(g(x_i)) - f_s(g(x_i))\rVert_{p} \\
 &+ \beta_2 \displaystyle \mathop{\mathbb{E}}_{r_j \sim D_{m} } \lVert f_w(r_j) - f_s(r_j)\rVert_{p},
\end{split}
\end{align}
where $\beta_1$ and $\beta_2$ are balancing weights. To mimic trial-to-trial variability in the brain, we inject noise into the logits of semantic memory ($\mathcal{\Tilde{S}}$) before applying consistency regularization as follows:
\begin{equation}
\label{eqn_logitn}
 f_s(r_j) \leftarrow f_s(r_j) + \delta
\end{equation}
where $\delta \sim \mathcal{N}(0, \sigma^2)$ is a white Gaussian noise, $\mathcal{L}_{cr}$ represents the expected Minkowski distance between the corresponding pairs of predictions and $p=2$. Consistency regularization enables the working model to retrieve structural knowledge from the semantic memory from previous tasks. Consequently, the working model adapts the decision boundary to new tasks without catastrophically forgetting previous tasks. 

Thus, the final learning objective for the working model is as follows:
\begin{equation}
\label{eqn_final}
 \mathcal{L} \triangleq \mathcal{L}_{repr} + \rho \;\mathcal{L}_{cr}
\end{equation}
where $\rho$ is a balancing parameter. Our proposed approach is illustrated in Figure \ref{fig:representation_replay} and is detailed in Algorithm \ref{alg:method}. 

Note that these noises are applied stochastically, and therefore, a single representation can have multiple noises associated with it. Although noise is generally treated as a nuisance, BiRT introduces controllable noise at various stages of the vision transformer to promote robust generalization in CL.

\begin{table*}[t]
\centering
\caption{Results on CIFAR-100 learned with 5, 10, and 20 tasks with varying buffer sizes. BiRT achieves consistent improvements over the state-of-the-art on average accuracy and last accuracy.}
\label{tab:sota_comparison}
\vskip 0.05in 
\begin{small}
\begin{sc}
\begin{tabular}{l|c|c|cc|cc|cc}
\toprule
\multirow{2}{*}{Methods} & \multirow{2}{*}{\begin{tabular}[c]{@{}c@{}}Buffer\\Size\end{tabular}} & \multirow{2}{*}{\#P} & \multicolumn{2}{|c}{5 steps} & \multicolumn{2}{|c}{10 steps} & \multicolumn{2}{|c}{20 steps} \\
 &  &  & {Avg} & {Last} & {Avg} & {Last} & {Avg} & {Last} \\
\midrule
ATT-asym  & - & 16.87 & - & - & 25.58 \tiny{$\pm$0.01} & 16.31 \tiny{$\pm$0.00} & - & - \\
FUNC-asym & - & 16.87 & - & - & 25.95 \tiny{$\pm$0.00} & 16.21 \tiny{$\pm$0.01} & - & - \\
BiRT      & - & 10.73 & - & - & \textbf{56.40} \tiny{$\pm$1.57} & \textbf{42.59} \tiny{$\pm$0.84} & - & - \\ 
 \midrule
DyTox & & 10.73 & 56.98 \tiny{$\pm$0.61} & 41.50 \tiny{$\pm$1.00} & 48.31 \tiny{$\pm$1.23} & 23.92 \tiny{$\pm$1.11} & 38.10 \tiny{$\pm$1.72} & 14.27 \tiny{$\pm$0.94} \\
LVT & \multirow{1}{*}{200} & 8.9 & - & 39.68 \tiny{$\pm$1.36} & - & 35.41 \tiny{$\pm$1.28} & - & 20.63 \tiny{$\pm$1.14} \\
BiRT &  & 10.73 & \textbf{67.15} \tiny{$\pm$0.95} & \textbf{54.15} \tiny{$\pm$0.94} & \textbf{61.01} \tiny{$\pm$1.58} & \textbf{45.59} \tiny{$\pm$1.54} & \textbf{48.03} \tiny{$\pm$0.97} & \textbf{29.10} \tiny{$\pm$1.88} \\ 
\midrule
DyTox &  & 10.73 & 63.85 \tiny{$\pm$0.99} & 52.99 \tiny{$\pm$0.53} & 58.35 \tiny{$\pm$1.54} & 34.54 \tiny{$\pm$1.82} & 49.98 \tiny{$\pm$1.32} & 24.86 \tiny{$\pm$0.81} \\
LVT & \multirow{1}{*}{500} & 8.9 & - & 44.73 \tiny{$\pm$1.19} & - & 43.51 \tiny{$\pm$1.06} & - & 26.75 \tiny{$\pm$1.29} \\
BiRT &  & 10.73 & \textbf{68.40} \tiny{$\pm$1.56} & \textbf{55.65} \tiny{$\pm$0.99} & \textbf{63.82} \tiny{$\pm$1.80} & \textbf{50.20} \tiny{$\pm$0.67} & \textbf{50.34} \tiny{$\pm$1.64} & \textbf{30.22} \tiny{$\pm$1.63} \\
\bottomrule
\end{tabular}
\end{sc}
\end{small}
\vskip -0.1in
\end{table*}

\begin{figure*}[tb]
\vskip 0.2in
\begin{center}
\centerline{
\begin{tabular}{ccc}
\includegraphics[width=.335\textwidth]{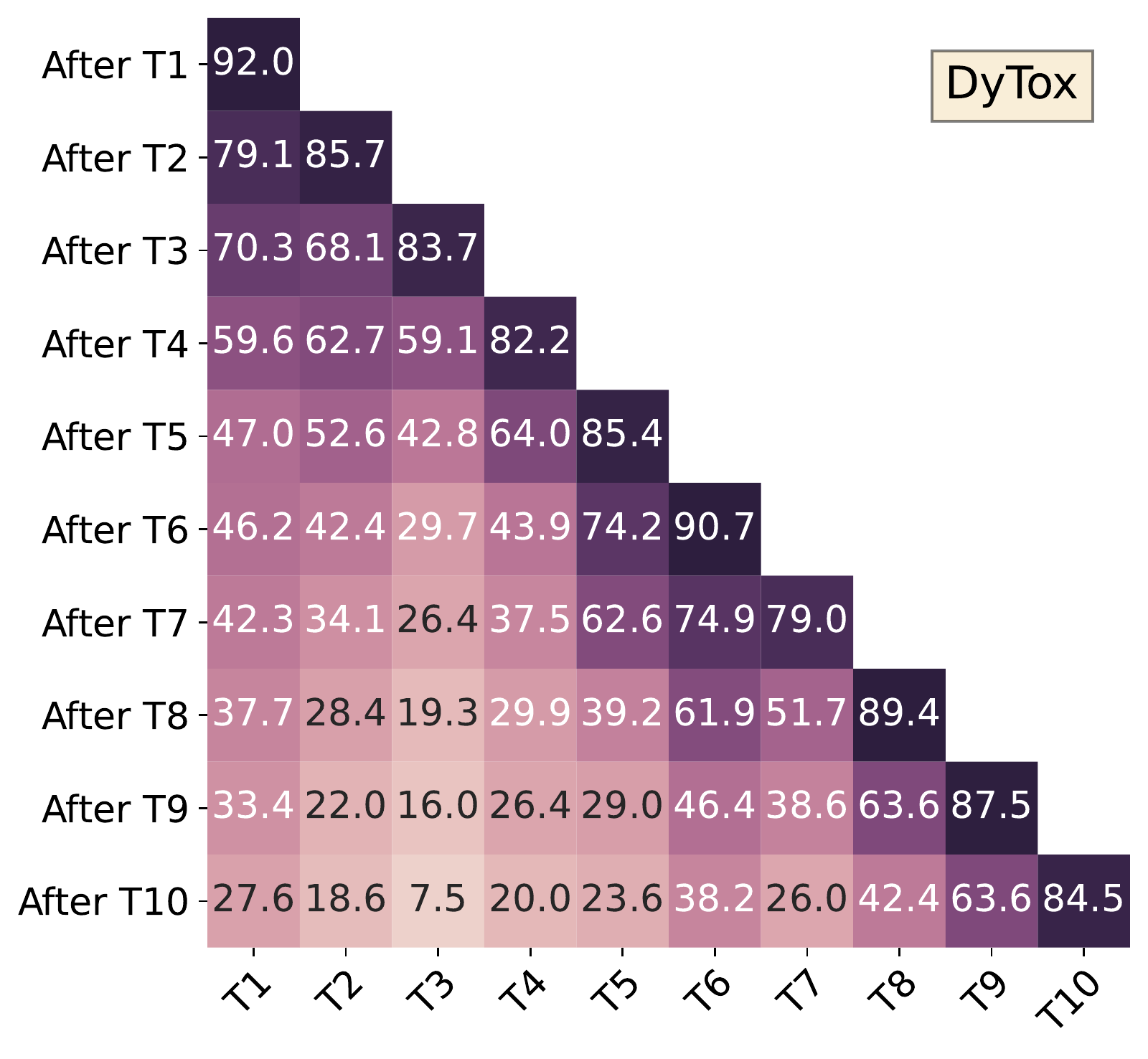} &  \includegraphics[width=.3\textwidth]{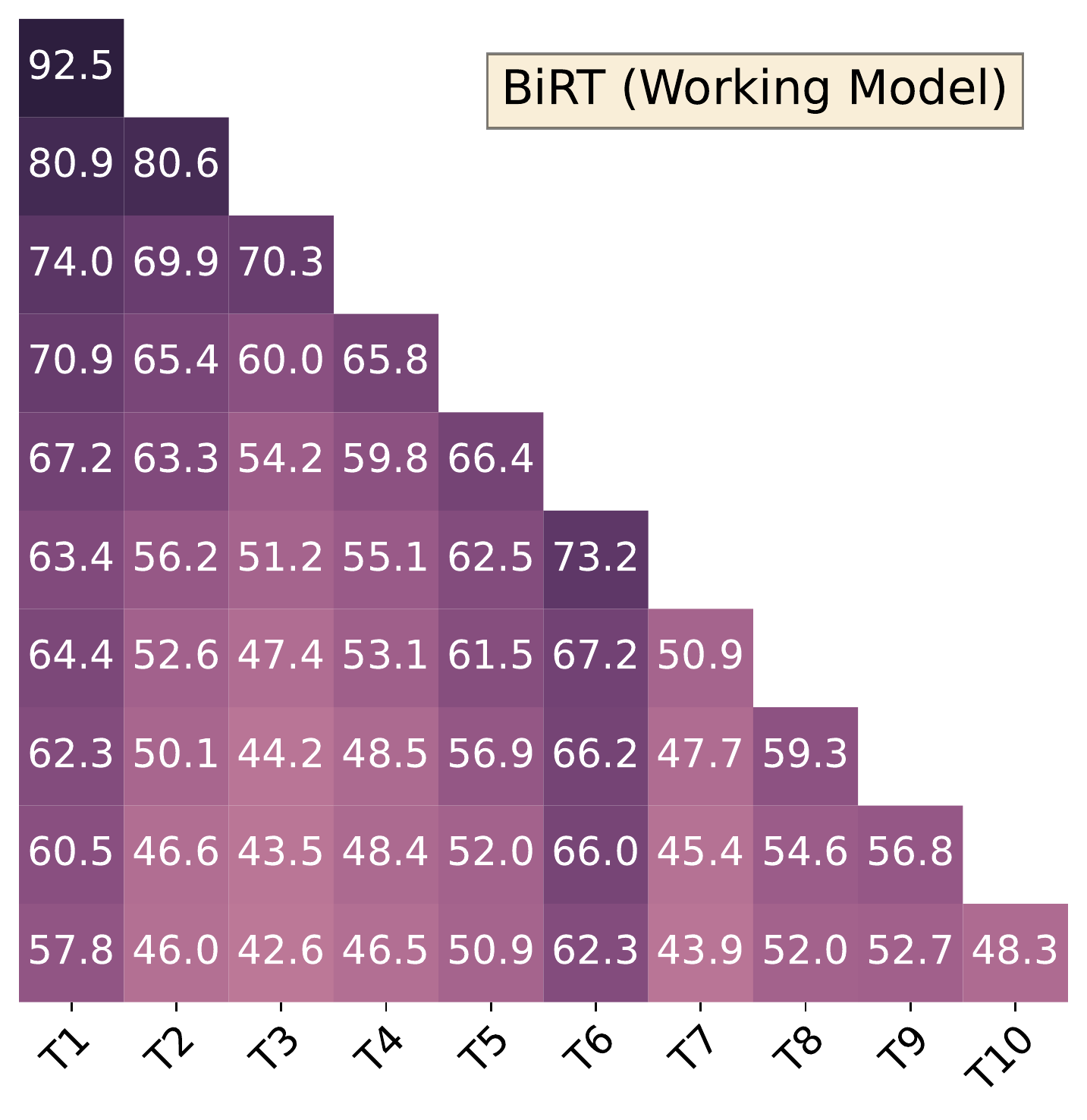} &
\includegraphics[width=.3\textwidth]{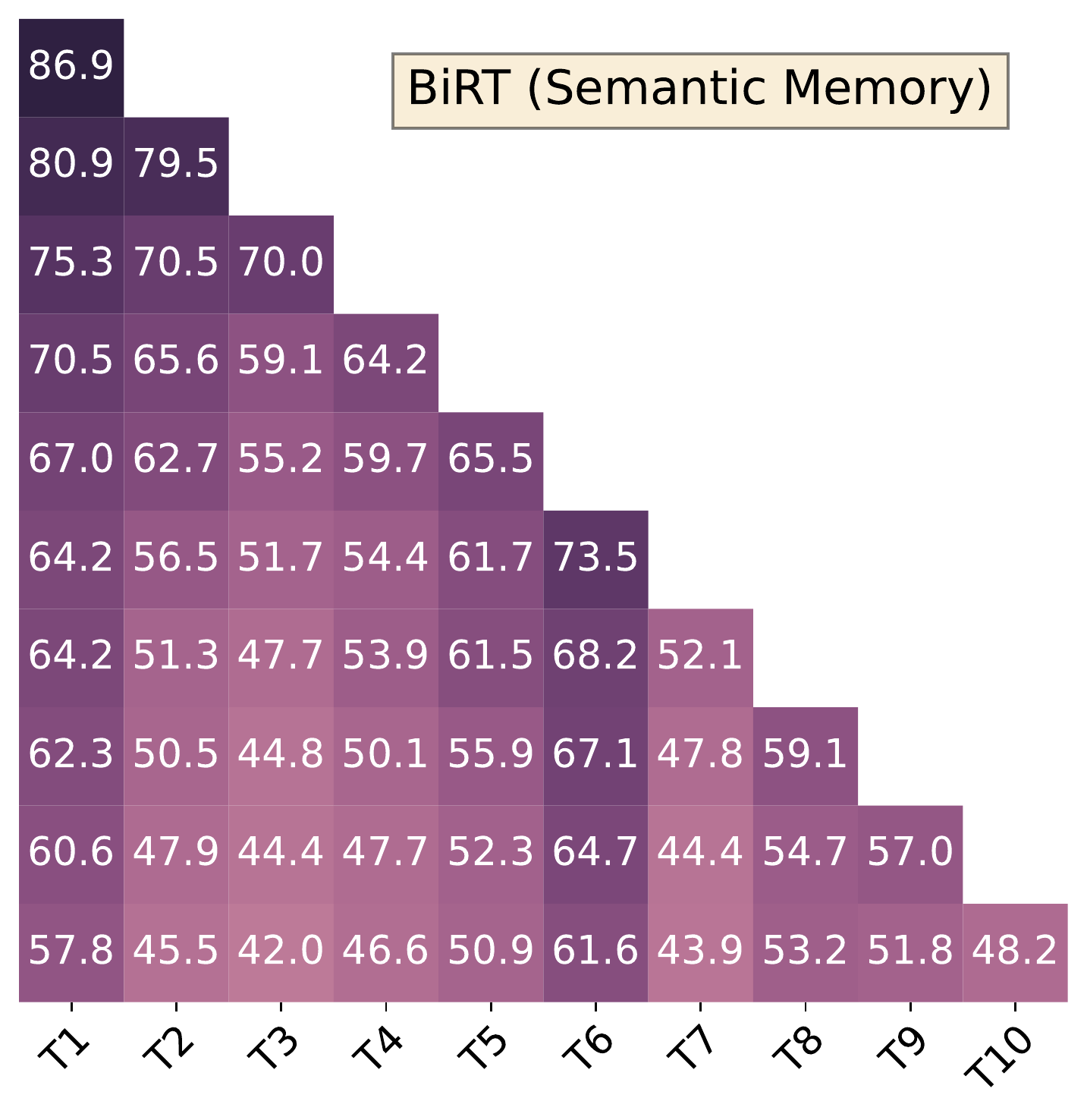}
\end{tabular}}
\caption{Comparison of task-wise performance after learning each task on CIFAR-100 with a buffer size of 500 learned for 10 tasks. The working model achieves better accuracy for the seen tasks after learning 10 tasks compared to DyTox. The semantic memory retains the performance of older tasks better than DyTox and the working model.}
\label{fig:taskwise_perfromance}
\end{center}
\vskip -0.2in
\end{figure*}

\section{Experimental Results}
We use the continuum library \citep{douillardlesort2021continuum} to implement different CL scenarios and build our approach on top of DyTox \citep{douillard2021dytox} method, the main baseline in all our experiments. We report the last accuracy (Last), average accuracy (Avg), forward transfer (FWT), backward transfer (BWT) and forgetting. More information on experimental setup, datasets, and metrics can be found in Appendix \ref{sec:exp_setup}.

Table \ref{tab:main} presents the comparison of our method with standard CL benchmarks with different buffer sizes, averaged across three random seeds. 
We can make the following observations from Table \ref{tab:main}: (i) Across CL settings and different buffer sizes, BiRT shows consistent performance improvement over DyTox across all metrics. 
(ii) BiRT enables the consolidation of rich information about the previous tasks better even under low buffer regimes, e.g. for CIFAR-100, the absolute improvement in terms of Last Acc is $7.28\%$ for buffer size 1000 while it is as much as $15.66\%$ for buffer size 500. (iii) BWT and FWT elucidate the influence of learning a new task $t$ on the performance of previous and subsequent tasks, respectively. 
BiRT shows a smaller negative BWT and a higher positive FWT across all CL datasets, resulting in less forgetting and better forward facilitation. (iv) TinyImageNet is one of the challenging datasets for CL considered in this work. Under low buffer regimes, the number of samples per class will be severely limited due to the large number of classes per task. BiRT consistently outperforms DyTox across all buffer sizes on TinyImageNet. 

Table \ref{tab:sota_comparison} further demonstrates the comparison of our method with transformer-based exemplar-free (ATT-asym and FUNC-asym \citep{pelosin2022towards}; averaged over 3 seeds) and rehearsal-based (DyToX and LVT; averaged over 5 class orderings) approaches. 
Although originally not designed for the exemplar-free scenario, BiRT shows a significant improvement over the rehearsal-free methods. 
Progressing from the exemplar-free scenario, BiRT shows a further improvement in performance when provided with experience rehearsal.
We also compare CL methods with different numbers of tasks in CIFAR-100 with limited buffer sizes. BiRT consolidates generalizable features rather than discriminative features specific to buffered samples, thereby exhibiting superior performance across all buffer sizes and task sequences.

\begin{figure*}[tb]
\vskip 0.2in
\begin{center}
\centerline{
\begin{tabular}{cc}
\includegraphics[width=0.23\textwidth]{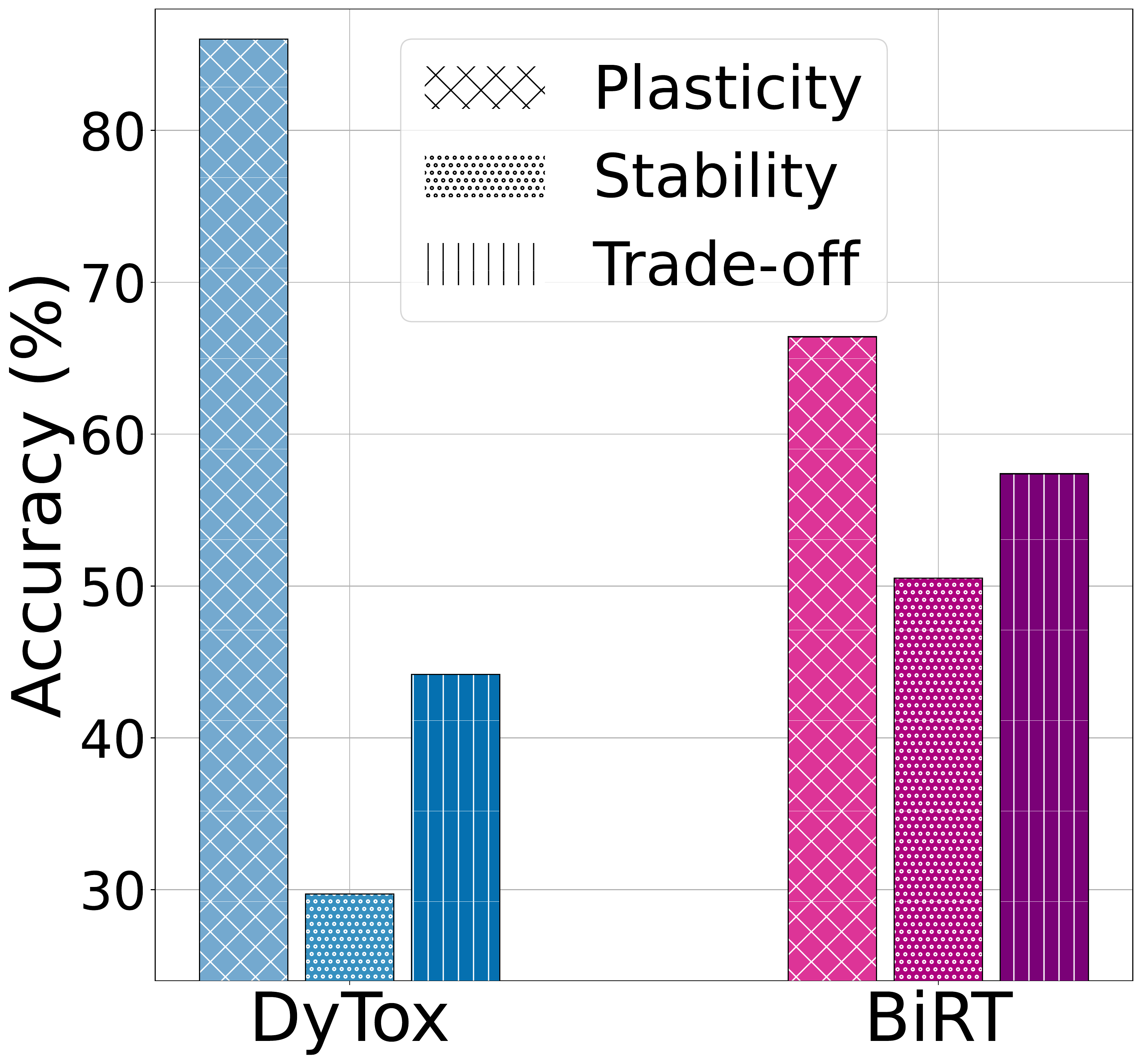} & 
\includegraphics[width=0.595\textwidth]{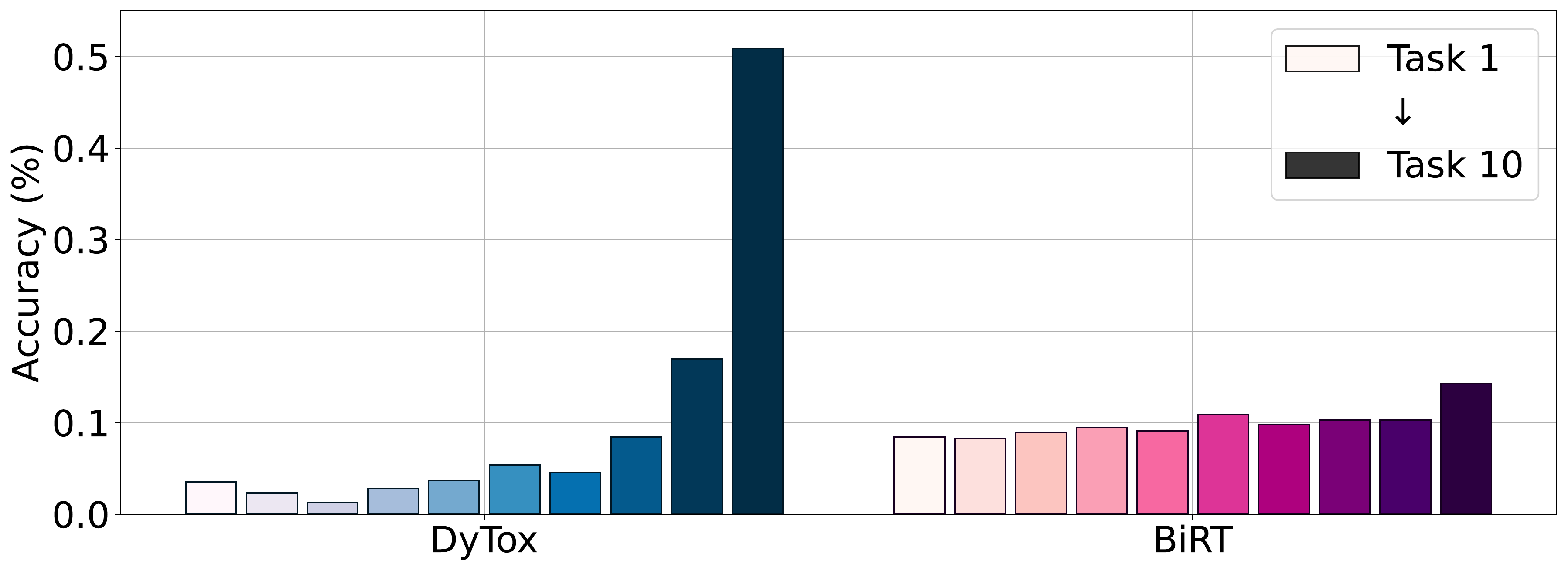}
\end{tabular}}
\caption{Comparison of CL methods in the stability-plasticity trade-off (left) and the task-recency bias (right) on C-100 (buffer size 500).}
\label{fig:tradeoff_recency}
\end{center}
\vskip -0.2in
\end{figure*}

Reinforcing our earlier hypothesis, the controllable noises introduced in BiRT play a constructive role in promoting generalization and consequently reducing overfitting in CL. In addition to allaying privacy concerns, replacing raw image rehearsal with representation rehearsal reduces the memory footprint without compromising performance.

\begin{figure*}[tb]
\vskip 0.2in
\begin{center}
\centerline{
\includegraphics[width=\linewidth]{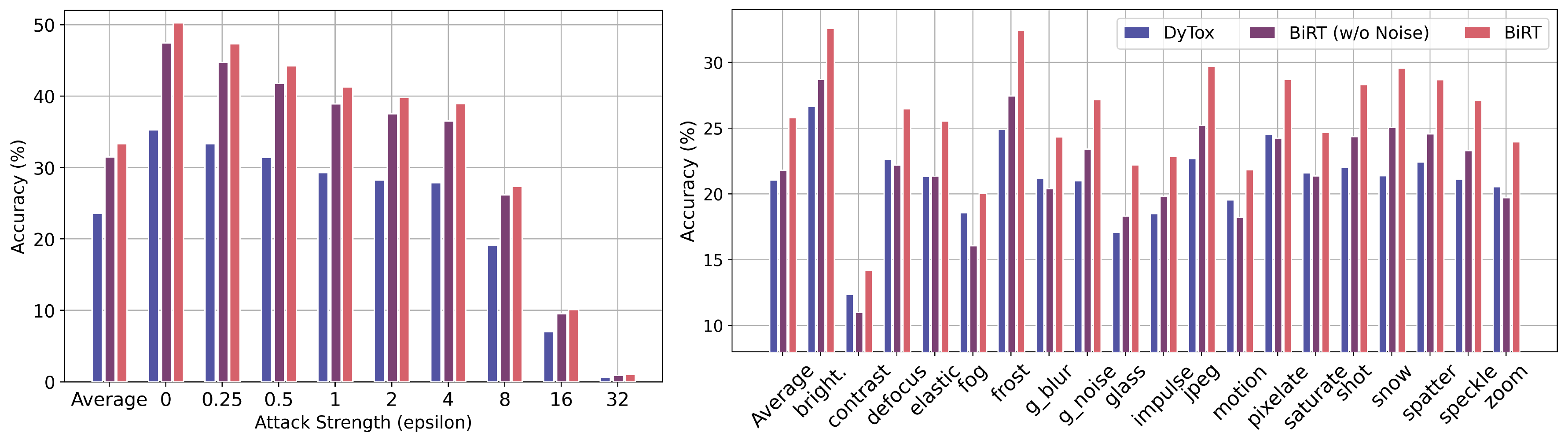}}
\caption{Robustness of CL methods to adversarial attacks (left) and 19 different natural corruptions (right) on C-100 (buffer size 500).}
\label{fig:robustness}
\end{center}
\vskip -0.2in
\end{figure*}

\section{Model Analysis}

\textbf{Task Recency Bias:}
Sequential learning of multiple tasks causes classifier predictions to tilt toward recent tasks, resulting in a task recency bias \citep{masana2020class}. One direct consequence of task recency bias is that the classifier norm is higher for recent classes while lower for older classes, which means that older classes are less likely to be picked for prediction \citep{hou2019learning}. 
Following the analysis in \citep{bhat},  Figure \ref{fig:tradeoff_recency} (right) shows the normalized probability that all classes in each task are predicted at the end of training. The probabilities in BiRT are more evenly distributed than in DyTox, resulting in a lower recency bias. We argue that supervision noises proposed in BiRT implicitly regularize the classifier towards more evenly distributed prediction probabilities.

\textbf{Stability-Plasticity Dilemma:}
The extent to which the CL model is plastic enough to acquire new information while stable enough not to catastrophically interfere with consolidated knowledge is referred to as stability-plasticity dilemma \citep{parisi2019continual}. Catastrophic forgetting is a direct consequence of this dilemma when the plasticity of the CL model overtakes its stability. To investigate how well our method handles the stability-plasticity dilemma, we plot the task-wise performance at the end of each task in Figure \ref{fig:taskwise_perfromance} for the CIFAR-100 test set. Following \citet{sarfraz2022synergy}, we also visualize a formal trade-off measure in Figure \ref{fig:tradeoff_recency} (left). Both the working model and semantic memory exhibit higher stability, while DyTox is more plastic. Therefore, DyTox is more prone to forgetting, whereas BiRT displays a better stability-plasticity trade-off compared to the baseline.

\begin{figure}[t]
\vskip 0.2in
\begin{center}
\centerline{\includegraphics[width=0.86\linewidth]{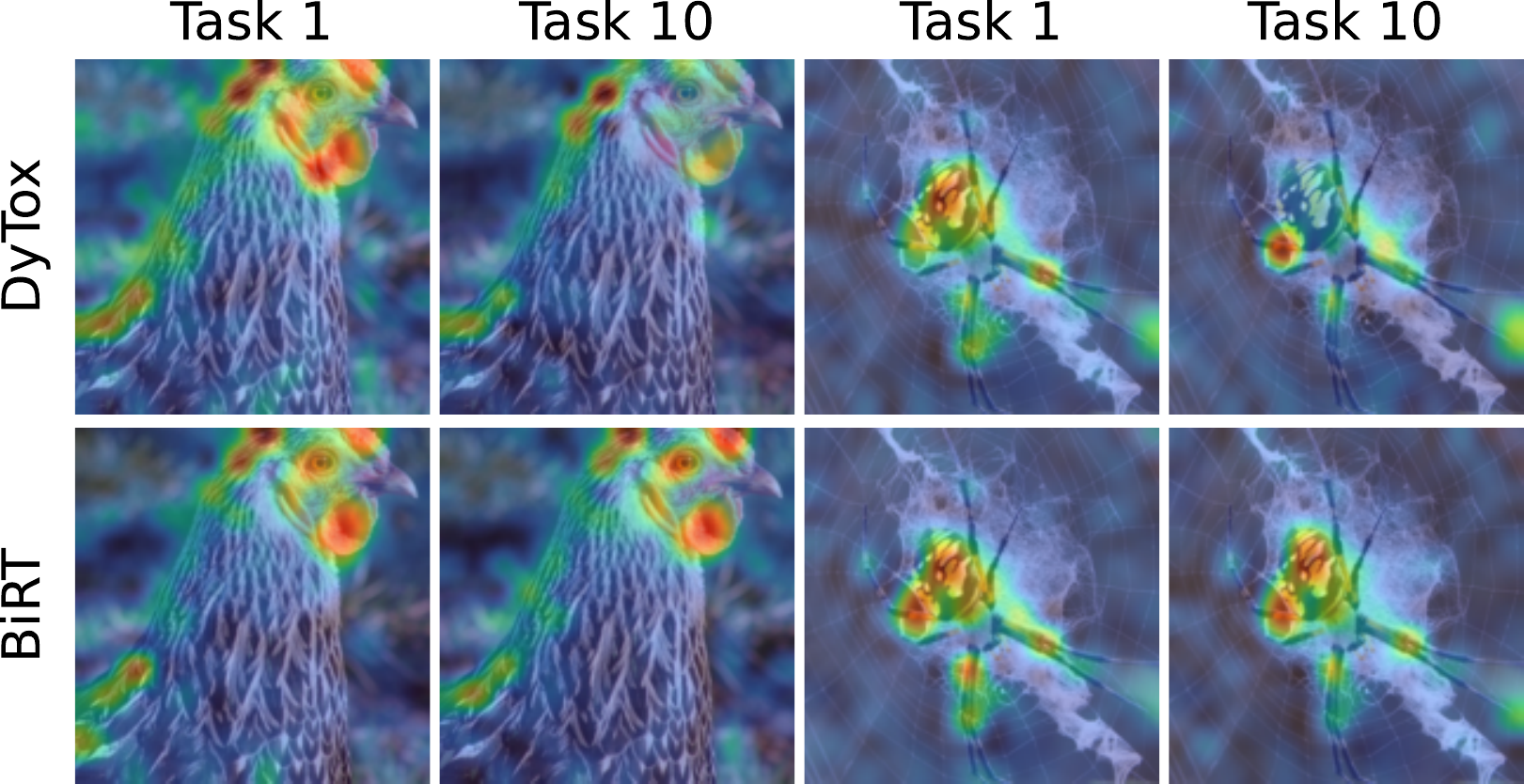}}
  \caption{Comparison of attention maps on the validation set of the first task of ImageNet-100 trained for 10 tasks with buffer size 500 (red regions correspond to regions with higher attention). BiRT retains the knowledge of salient regions in the image better than DyTox, leading to better predictions and less forgetting.}
  \label{fig:short_attention}
\end{center}
\vskip -0.2in
\end{figure}

\textbf{Attention Map Analysis:}
As learning progresses through a sequence of tasks, a CL model that retains its focus on salient regions undergoes less catastrophic forgetting. Therefore, it would be beneficial to study the variation in the salient regions of the image during the learning trajectory. Figure \ref{fig:short_attention} shows a comparison of saliency maps for samples of the first task after training on the first and last task, respectively. As can be seen, BiRT retains the attention to important regions in these images better than DyTox. 
We contend that the attention noise proposed in BiRT helps focus on class-wide features rather than sample specific features, thereby retaining attention to important regions in test images. 
More explanation and extended visualizations are provided in Appendix \ref{attention_maps}. 

\textbf{Robustness Analysis:}
Continual learning models are mostly evaluated on accuracy on seen tasks and forgetting metrics. 
However, the research community has largely neglected the susceptibility of continually learned models to adversarial attacks and corrupted data in the wild \citep{khan2022susceptibility}. 
Figure \ref{fig:robustness} illustrates the robustness of BiRT on adversarial attack of varying strengths \citep{kim2020torchattacks} and several natural corruptions \citep{hendrycks2019benchmarking}. 
In addition, we evaluate the robustness of BiRT without any noise in the learning trajectory in order to elucidate the benefits of constructively inducing noise in the pipeline of continually learning models. 
BiRT is robust to adversarial attacks, as well as corrupted data, and learning with noise results in improved robustness. 
This is evident from the performance under severe noises such as \emph{`contrast'}, \emph{`fog'}, \emph{`motion blur}' and the average performance across different settings wherein learning with noise helps the model recover from the inferior performance.

This makes it well-suited for safety-critical applications, such as autonomous vehicles, where the consequences of a model failure can be severe.

\section{Ablation Study}

Table \ref{tab:birt_ablation} provides an overview of the effect of the different components used in BiRT. Unlike DyTox, we employ an exponential moving average as semantic memory, resulting in the biggest jump in accuracy.  BiRT entails representation, attention, and supervision noises to promote robust generalization in CL and diversify the buffered representations. As can be seen, all three components of BiRT play a constructive role in building a successful continual learner. 
Supervision noise, representation noise, and attention noise bring performance improvements of 0.54\%, 3.30\%, and 3.41\%, respectively, over BiRT without any noise.
In addition, compared to vanilla representation rehearsal, the right combination of controllable noises in BiRT greatly reduces overfitting and improves performance by as much as $9\%$ (relative Avg). 
Therefore, it is quintessential to have controllable noise to further improve representation rehearsal in CL.

\begin{table}[t]
\centering
\caption{Ablations of the different key components of BiRT. The average and last accuracies are reported on CIFAR100 for the buffer size of 500 learned for 10 tasks.}
\label{tab:birt_ablation}
\vskip 0.05in
\begin{small}
\begin{sc}
\resizebox{\columnwidth}{!}{
\begin{tabular}{@{}ccc|c|c@{}}
\toprule
\begin{tabular}[c]{@{}c@{}}Supervision \\ Noise\end{tabular} & \begin{tabular}[c]{@{}c@{}}Repres. \\ Noise\end{tabular} & \begin{tabular}[c]{@{}c@{}}Attention \\ Noise\end{tabular} &
\begin{tabular}[|c]{@{}c@{}}Last\\ Acc\end{tabular} &
\begin{tabular}[|c]{@{}c@{}}Avg\\ Acc\end{tabular} \\
\midrule
\cmark & \cmark & \cmark & \textbf{50.20}\tiny{$\pm$0.67} & \textbf{63.82}\tiny{$\pm$1.80} \\
\xmark & \cmark & \cmark & 49.63\tiny{$\pm$0.30} & 63.67\tiny{$\pm$1.55} \\
\xmark & \xmark & \cmark & 49.30\tiny{$\pm$0.91} & 63.29\tiny{$\pm$1.71} \\
\cmark & \xmark & \xmark & 49.19\tiny{$\pm$0.46} & 62.58\tiny{$\pm$1.44} \\
\xmark & \cmark & \xmark & 46.43\tiny{$\pm$0.41} & 61.83\tiny{$\pm$0.23} \\
\xmark & \xmark & \xmark & 45.89\tiny{$\pm$1.25} & 59.58\tiny{$\pm$0.58} \\ \midrule
\multicolumn{3}{c|}{DyTox} & 34.54\tiny{$\pm$1.82} & 58.35\tiny{$\pm$1.54} \\
 \bottomrule
\end{tabular}}
\end{sc}
\end{small}
\vskip -0.1in
\end{table}

\section{Conclusions and Future Work}

We proposed BiRT, a novel representation rehearsal-based continual learning approach based on vision transformers. Specifically, we introduce controllable noises at various stages of the vision transformer
and enforce consistency in predictions with respect to an exponential moving average of the working model. Our empirical results show that BiRT outperforms raw image rehearsal and vanilla representation rehearsal while being memory efficient and robust to natural and adversarial corruptions. Furthermore, the improvement is even more pronounced under low buffer regimes and longer task sequences. Reinforcing our earlier hypothesis, the controllable noises introduced in BiRT play a constructive role in promoting
generalization and consequently reducing overfitting in CL. 
Extending our work to more realistic settings such as general CL where task boundaries are not known at training time, and exploring other efficient transformer architectures are some of the useful research directions for this work.

\bibliographystyle{apalike}
\bibliography{ref.bib}

\newpage
\appendix
\onecolumn

\section{Experimental setup, datasets and metrics} 
\label{sec:exp_setup}

We use the continuum library \citep{douillardlesort2021continuum} to implement different CL scenarios and build our approach on top of DyTox \citep{douillard2021dytox} framework, which is the main baseline in all our experiments. 
We use a network that consists of 5 self-attention blocks and a task-attention block. All blocks have 12 attention heads and an embedding dimension of 384. We train models with a learning rate of $5e^{-4}$, a batch size of 128, and a weight decay of $1e^{-6}$. All models, including the baseline, are trained for 500 epochs per task in CIFAR-100 \citep{krizhevsky2009learning}, TinyImageNet \citep{le2015tiny}, and ImageNet-100 \cite{deng2009imagenet}. During the patch embedding process, we utilize patch sizes of 4 for CIFAR-100, 8 for TinyImageNet, and 16 for ImageNet-100. After each task, the model is fine-tuned on a balanced dataset with a learning rate of $5e^{-5}$ for 20 epochs. All models are trained on a single NVIDIA V100 GPU, and all evaluations are performed on a single NVIDIA RTX 2080 Ti GPU.

We focus mainly on the class-incremental learning setting (Class-IL) \citep{van2019three}, where the task ID is not known at the test time. 
In every task, samples belonging to a new set of classes disjoint from the previous tasks' classes are learned by the model. 
Following \citet{douillard2021dytox} and \citet{de2021continual}, we evaluate our approach on CIFAR-100, ImageNet-100, and TinyImageNet.
CIFAR-100 consists of 50,000 training images and 10,000 test images of size 32$\times$32 belonging to 100 classes. ImageNet-100 consists of 129k train and 5,000 validation images of size 224$\times$224 belonging to 100 classes. TinyImageNet consists of 100,000 training images and 10,000 test images of size 96$\times$96 belonging to 200 classes.

Except for the analysis of longer task sequences, all other experiments are carried out in the Class-IL setting with 10 tasks.
In the case of CIFAR-100, 100 classes are divided into 10 tasks, with 10 classes in each task. 
Similarly, 20 classes per task are learned on TinyImageNet and 10 classes per task on ImageNet-100.
The order in which classes are learned can affect the performance of a CL model \citep{yoon2019scalable}. We use ``class order 1" from \citep{douillard2021dytox} for CIFAR-100 and ImageNet-100, and the sequential class order from 1 to 200 for TinyImageNet-200.

Although the performance of task-incremental learning (Task-IL) can be evaluated in our proposed approach, we exclude them in our analysis because it simplifies the CL scenario by assuming the availability of task id at the test time, which translates into choosing the right prediction head during inference. 

\subsection{Evaluation Metrics}

To evaluate the performance of different models under different settings, we select five different metrics widely used in the CL literature. We formalize each metric below.

\begin{enumerate}
 \item \textbf{Last Accuracy} \citep{douillard2021dytox} defines the final performance of the CL model on the validation set of all the tasks seen so far. Concretely, given that tasks are sampled from a set $t \in {1,2 ..., T}$, where $T$ is the total number of tasks and $a_{k,j}$ is the accuracy of a CL model on the validation set of the task $k$ after learning task {$j$}, last accuracy $A_{last}$ is as follows:
 \begin{equation}
A_{last} = \frac{1}{T} \sum_{k=1}^{T} a_{k, T}
 \end{equation}
 
 \item \textbf{Average Accuracy} \cite{rebuffi2017icarl} defines the average performance of the learned CL model on the validation set of all tasks seen so far after learning each task. Given that $K$ is the number of tasks seen so far and $T$ is the total number of tasks, the average accuracy $A_{avg}$ is as follows: 
 \begin{equation}
A_{avg} = \frac{1}{T} \sum_{j=1}^{T} \frac{1}{K} \sum_{k=1}^{K} a_{k, j}
 \end{equation}
 
 \item \textbf{Backward Transfer (BWT)} \citep{lopez2017gradient} defines the influence of the learning task $t$ on previously seen tasks $k < t$. Positive BWT implies that the learning task $t$ increased performance on previous tasks, while negative BWT indicates that the learning task $t$ affected the performance of the model on previous tasks. Formally, BWT is as follows:
 \begin{equation}
BWT = \frac{1}{T-1} \sum_{j=1}^{T-1} a_{T, j} - a_{j, j}
 \end{equation}
 
 \item \textbf{Forward Transfer (FWT)} \citep{lopez2017gradient} defines the influence of learning the task $t$ on future tasks $k > t$. Positive FWT implies that learning the task $t$ increased performance in future tasks and vice versa. Positive FWT occurs when the model learns generalizable features that can help it learn future tasks. Formally, given that $\hat{a}_j$ is the accuracy of the task $j$ at random initialization, the FWT is as follows:
 \begin{equation}
FWT = \frac{1}{T-1} \sum_{j=2}^{T} a_{j-1, j} - \hat{a}_{j}
 \end{equation}
 
 \item \textbf{Forgetting} \citep{chaudhry2018riemannian} quantifies the forgetting of previously learned tasks given the current state of the model. It is defined as the difference between the maximum accuracy of the model in previously learned tasks throughout the learning process and the current accuracy of the task. Concretely, forgetting for the task $k$ is as follows:
 \begin{equation}
\text{Forgetting} =\max_{l \in {1,2,...,k-1}} a_{k,l} - a_{k, T}
 \end{equation}
\end{enumerate}

\section{Quantitative results for figures}

To facilitate comparisons with BiRT, we provide quantitative results for the figures in the main text.

\subsection{Effect of Longer Sequences}
\begin{figure*}[t]
\vskip 0.2in
\begin{center}
\centerline{
\begin{tabular}{ccc}
     \includegraphics[width=.33\textwidth]{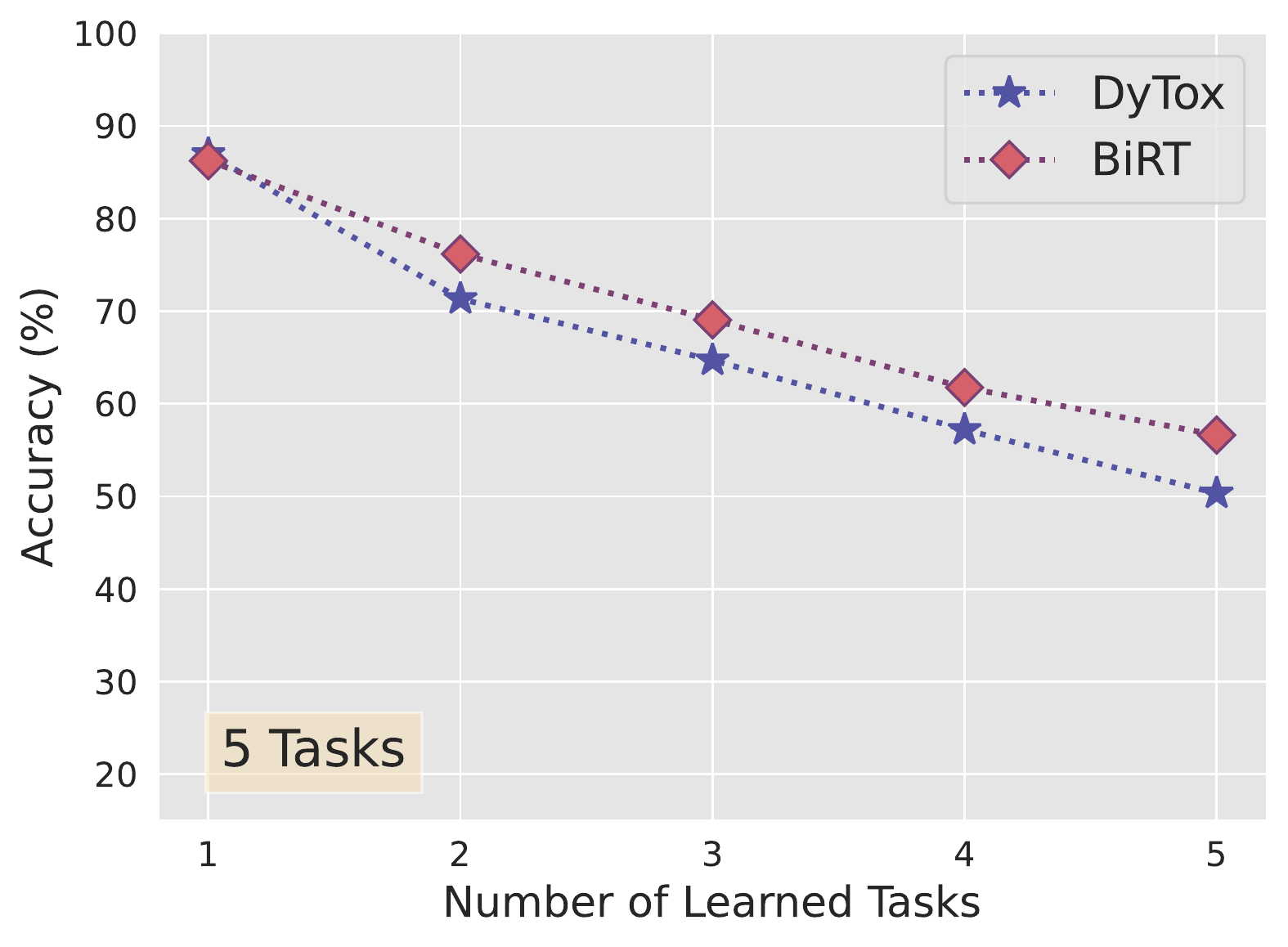} & 
     \includegraphics[width=.33\textwidth]{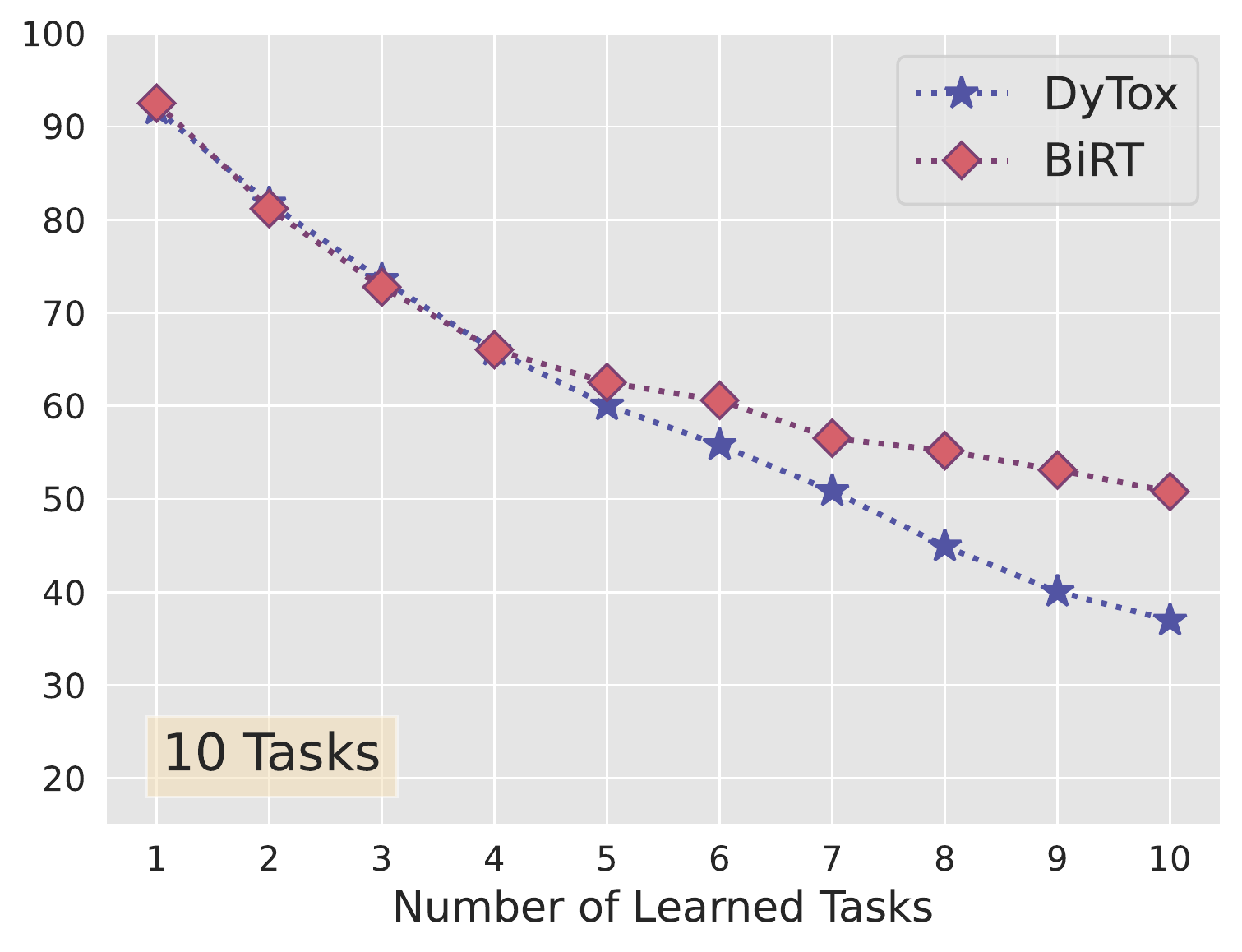} & 
     \includegraphics[width=.33\textwidth]{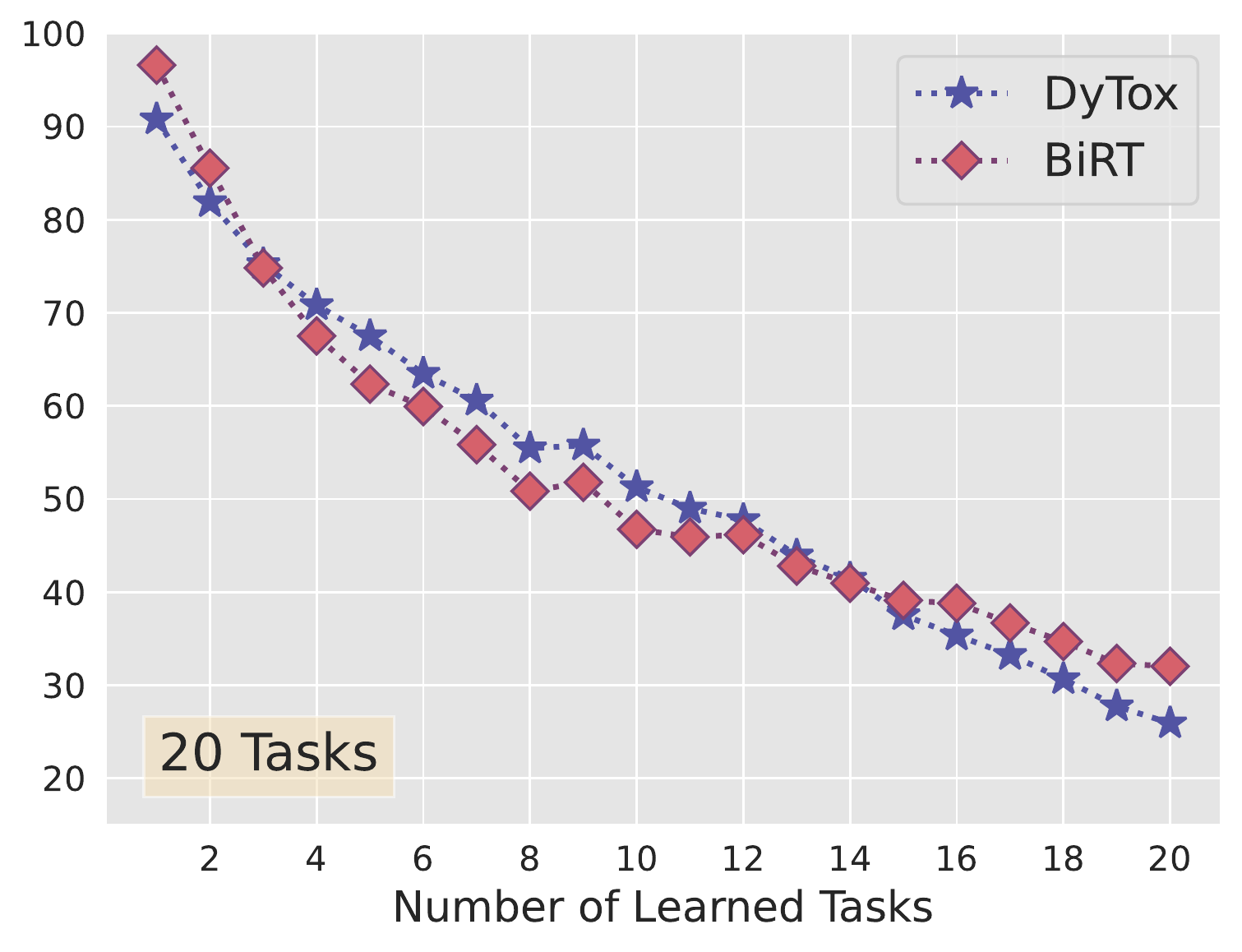} \\
\end{tabular}}
\caption{Comparison of performance across learning longer sequence of tasks on CIFAR-100.}
\label{fig:seq_tasks}
\end{center}
\vskip -0.2in
\end{figure*}

Given a limited buffer size, catastrophic forgetting worsens with increasing number of tasks, since the number of representative samples per task/class will be more limited \citep{peng2021overcoming}. To perform better, it is quintessential for the CL model to consolidate generalizable features rather than discriminative features specific to buffered samples. Figure \ref{fig:seq_tasks} presents the performance of CL models in sequences of 5, 10, and 20 tasks on CIFAR-100 with a buffer size of 500. Even as the number of tasks increases, BiRT maintains a substantial improvement over DyTox across all task sequences. As is the case with low-buffer regimes, BiRT consolidates the past task information better than the baseline, thereby further mitigating catastrophic forgetting.

\subsection{Stability-plasticity dilemma}

Figure \ref{fig:tradeoff_recency} (left) shows that BiRT achieves better stability, while DyTox is more plastic. We concluded that DyTox is more prone to forgetting, while BiRT exhibits a better stability-plasticity trade-off. We provide the numerical values for the same in Table \ref{tab:stability_numbers}. Note that the semantic memory of BiRT achieves a slightly higher stability-plasticity trade-off compared to the working model of BiRT (which is not clear in the illustration). 

\begin{table}[t]
\centering
\caption{Quantitative results for the stability-plasticity trade-off in CIFAR-100 for 10 tasks with buffer size 500.}
\label{tab:stability_numbers}
\vskip 0.05in
\begin{small}
\begin{sc}
\begin{tabular}{@{}lccc@{}}
\toprule
 & Plasticity & Stability & Trade-off \\ \midrule
DyTox & \textbf{86.06} & 29.74 & 44.16 \\
BiRT - Working Model & 66.42 & \textbf{50.52} & \textbf{57.38} \\
BiRT - Semantic Memory & 66.08 & 50.37 & 57.16 \\
\bottomrule
\end{tabular}
\end{sc}
\end{small}
\vskip -0.1in
\end{table}

\section{Working Principle of DyTox}

As mentioned in Section \ref{sec:exp_setup}, we build our proposed approach on top of DyTox framework \citep{douillard2021dytox}, an architecture expansion approach to continual learning with Transformers as the working model. DyTox uses the information about the task id during the training time to learn task-specific classifiers and task tokens. However, no task oracle is used during inference. 

DyTox architecture consists of 5 blocks of Self-Attention Blocks (SABs, implemented using ConVit \citep{DBLP:journals/corr/abs-2103-10697}) as an encoder to process the input image after the tokenization process. The features predicted by the encoder are then combined with a task token (which is specific to that task) and fed into a Task-Attention Block (TAB), in which the task token attends to the features and extracts the task-specific information. A task-specific classifier projects the processed task token to the number of classes in the task. Thus, the task token and classifier are expanded with respect to every task, while the SAB and TAB blocks are shared between tasks. Furthermore, it employs the copy of the working model at the task boundary as a teacher model to distill the information about past tasks into the working model. DyTox freezes the task tokens and classifier heads of previously learned tasks in order to retain the performance on old tasks.

\section{Model Analysis on Other Datasets}

We analyze task-wise probability (in Figure \ref{fig:taskwise_perfromance}), stability-plasticity trade-off, and task-recency bias (in Figure \ref{fig:tradeoff_recency}) on the CIFAR-100 dataset learned for 10 tasks with buffer size 500 in the main text. Here, we show additional results on other datasets (TinyImageNet and ImageNet-100). 

\begin{figure*}[t]
\begin{center}
\vskip 0.2in
\centerline{
 \begin{tabular}{ccc}
\includegraphics[width=.32\textwidth]{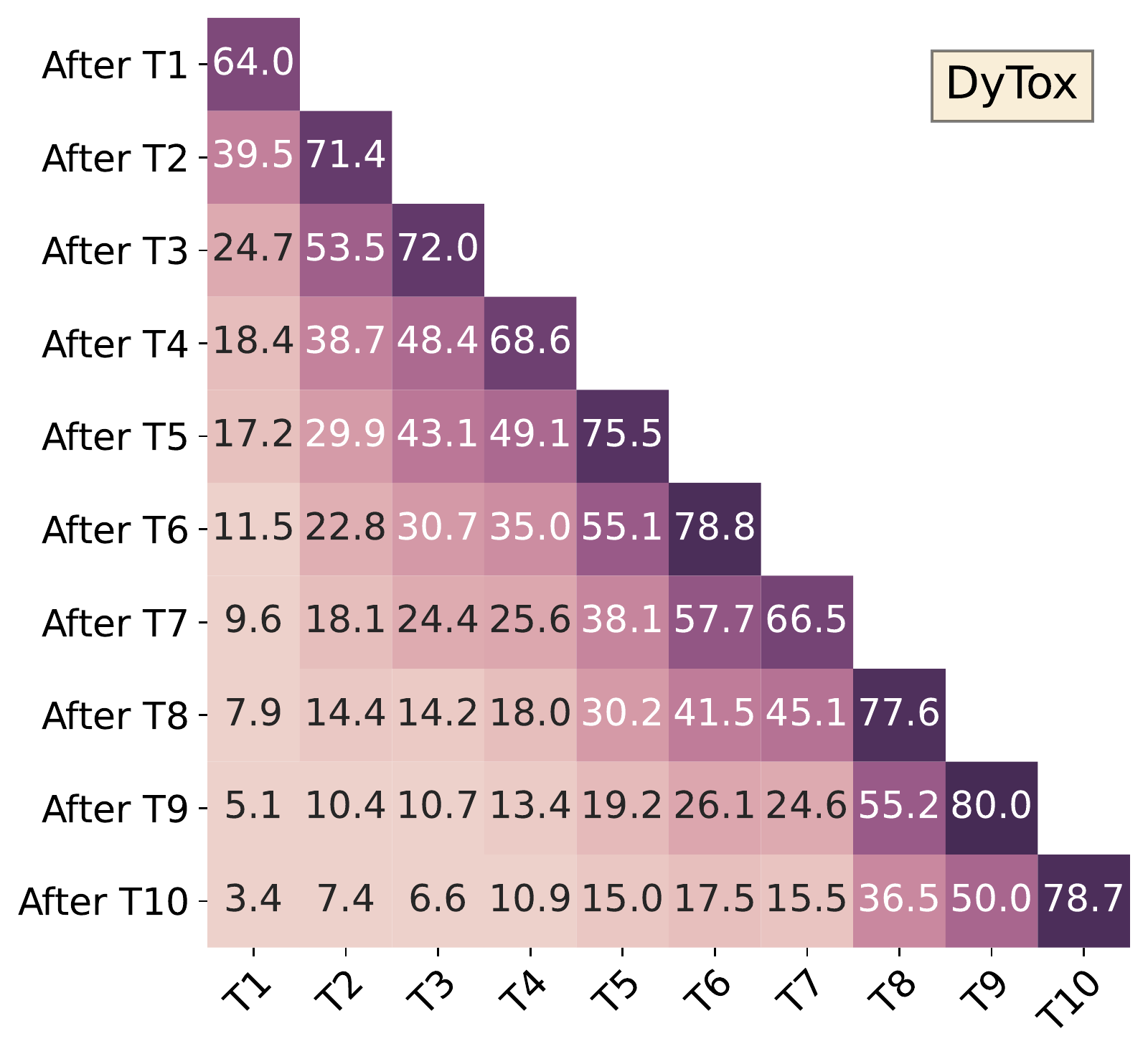} &  \includegraphics[width=.29\textwidth]{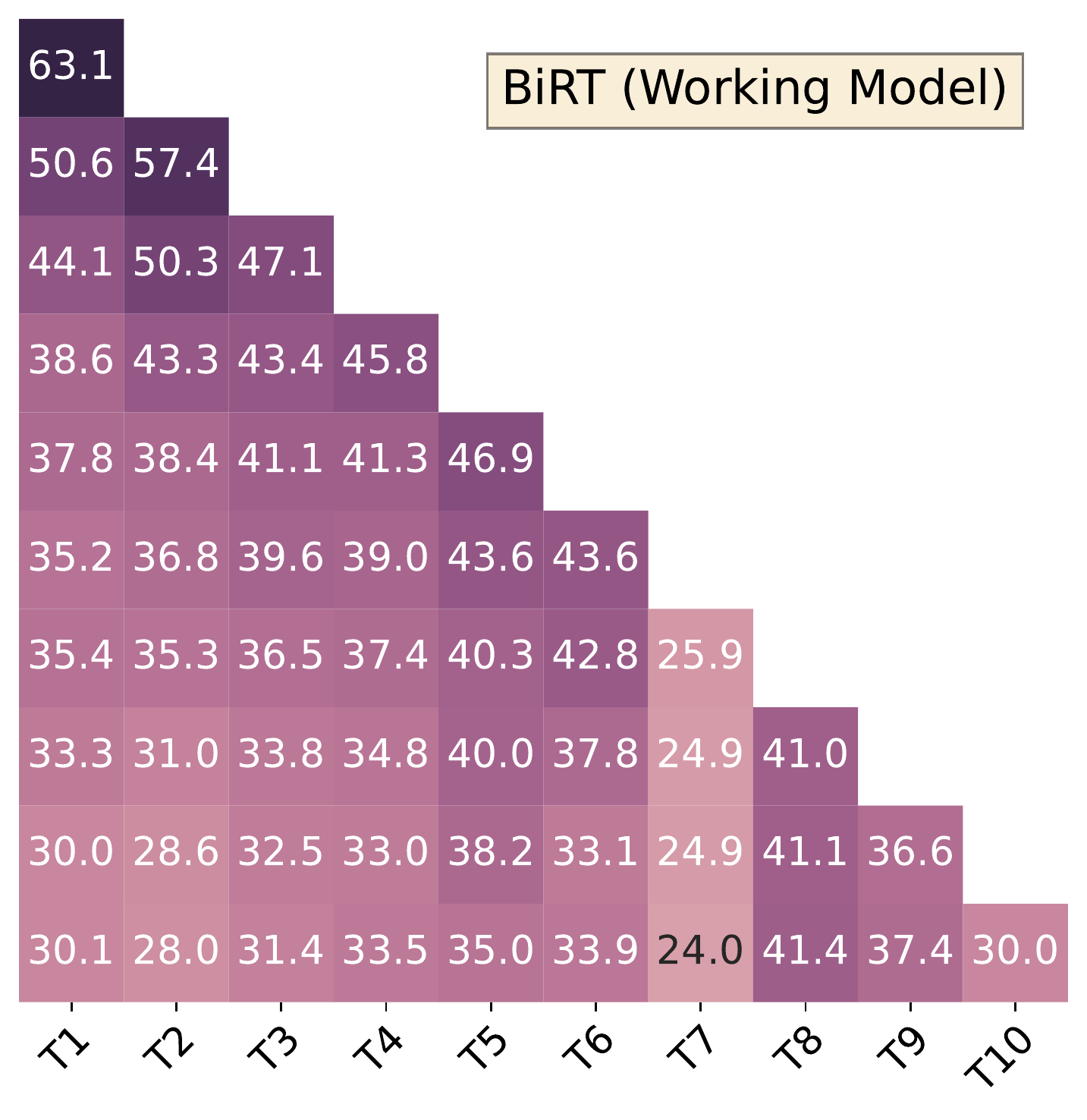} &
\includegraphics[width=.29\textwidth]{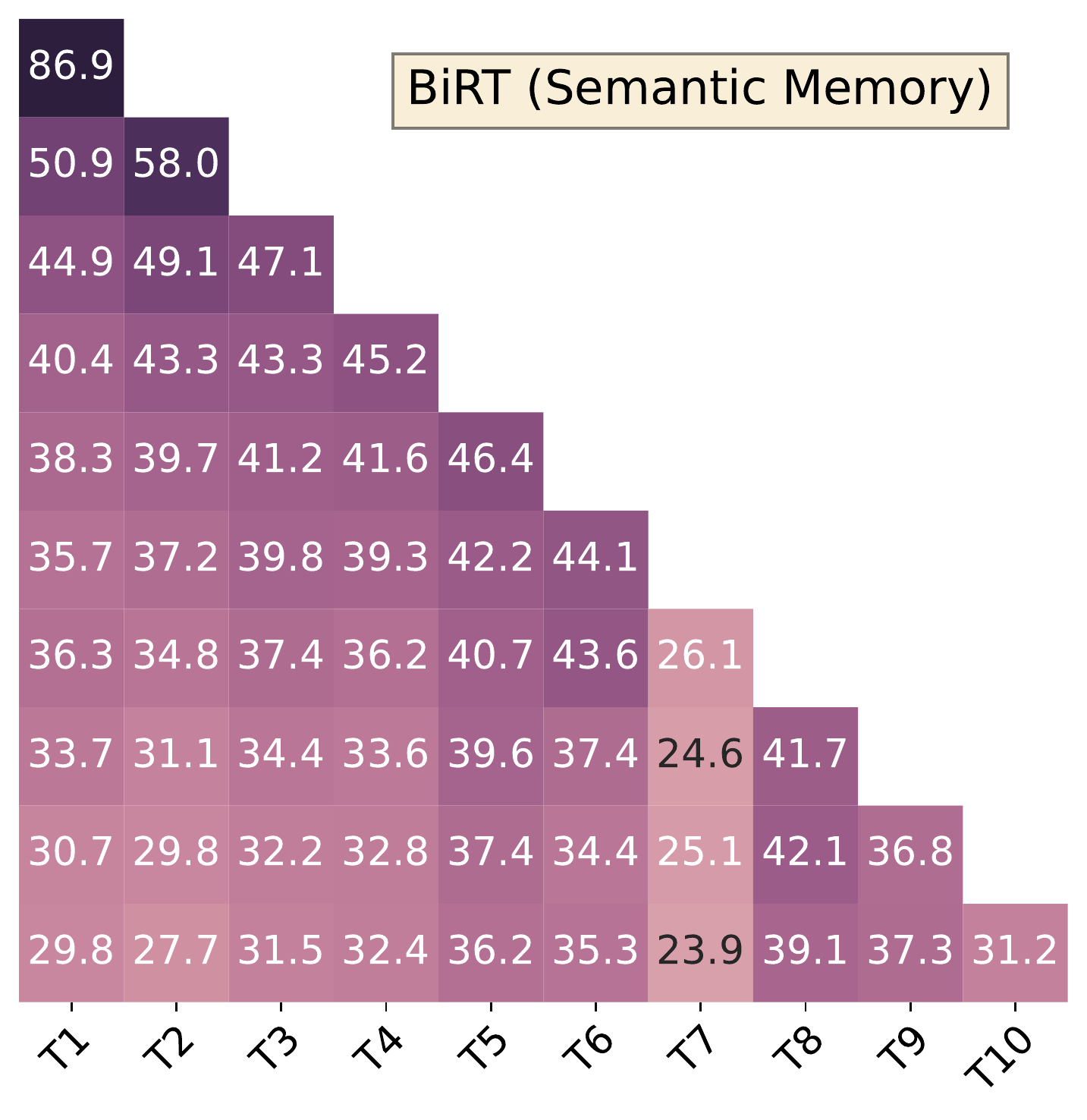}
 \end{tabular}}
\caption{Comparison of task-wise performance after learning every task on TinyImageNet with a buffer size of 500 learned for 10 tasks. The working model achieves better accuracy for the seen tasks after learning 10 tasks compared to DyTox. Semantic memory retains the performance of older tasks better than the baseline and working model.}
\label{fig:taskwise_perf_tiny}
\end{center}
\vskip -0.2in
\end{figure*}

\begin{figure*}[t]
\begin{center}
\vskip 0.2in
\centerline{
 \begin{tabular}{ccc}
    \includegraphics[width=.26\textwidth]{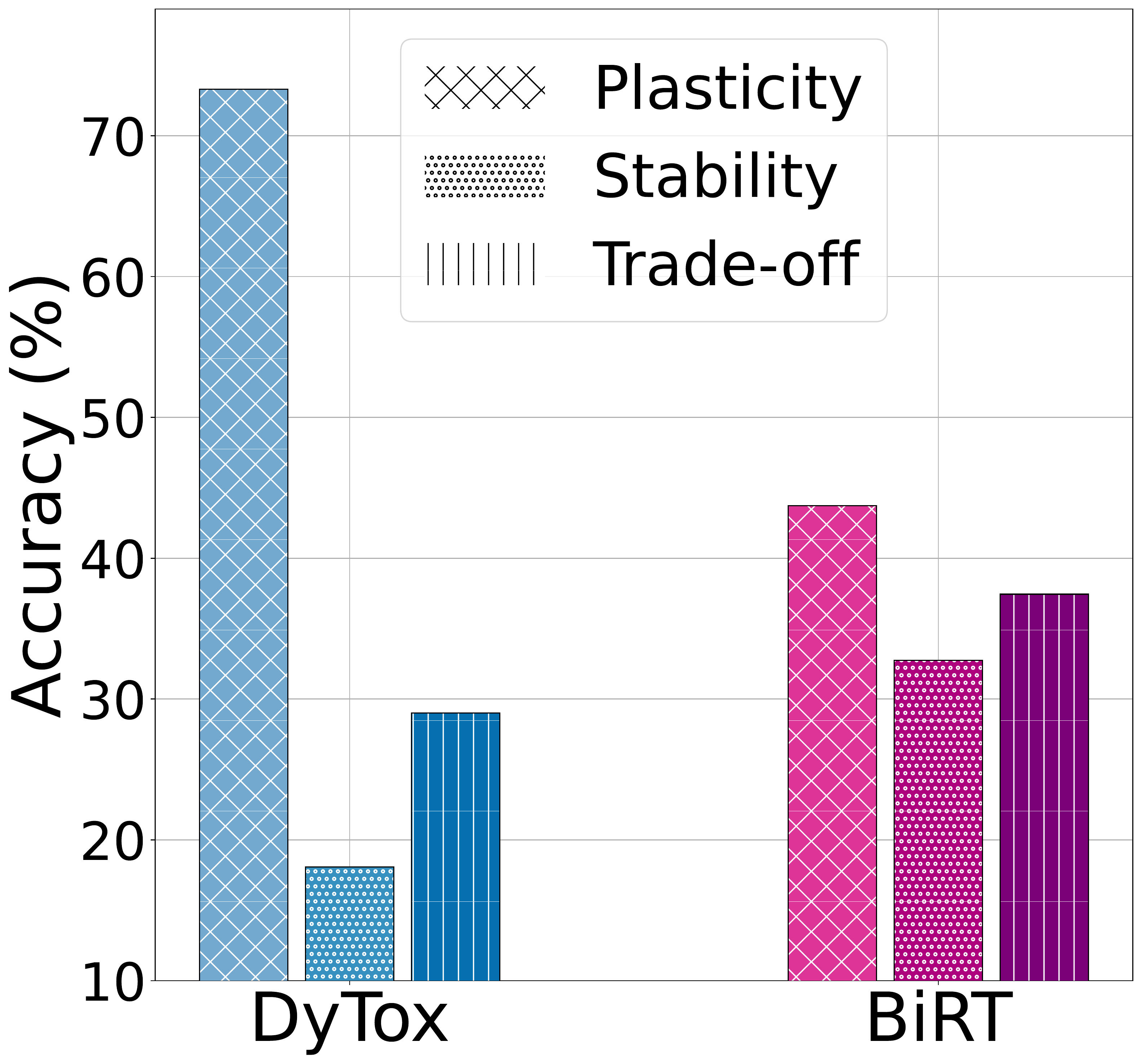} &
    \includegraphics[width=.67\textwidth]{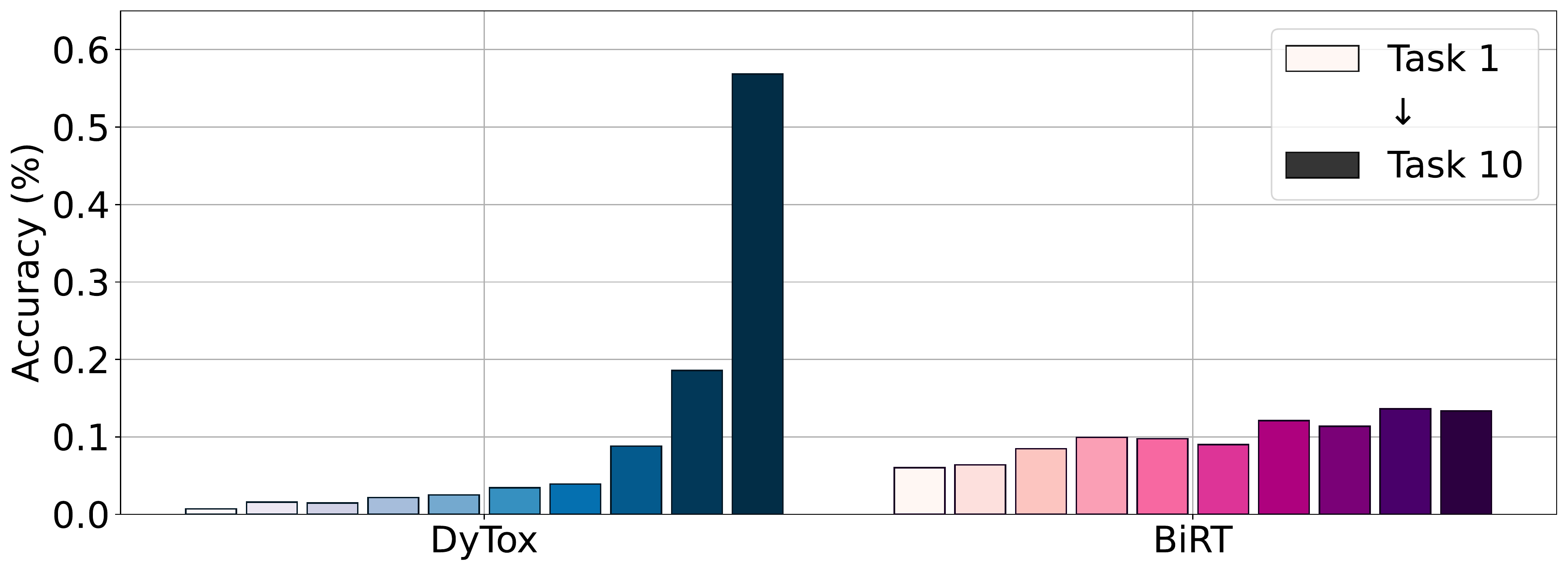} \\
\end{tabular}}
\caption{Comparison of the stability-plasticity trade-off (left) and the task-recency bias (right) trained for 10 tasks on TinyImageNet with buffer size 500.}
\label{fig:tradeoff_recency_tiny}
\end{center}
\vskip -0.2in
\end{figure*}

Figure \ref{fig:taskwise_perf_tiny} illustrates the task-wise accuracy of BiRT vs. DyTox in TinyImageNet. It is evident that BiRT (Semantic Memory) retains more knowledge about past tasks, which in turn helps BiRT (Working Model) achieve better overall performance compared to DyTox. The stability-plasticity trade-off shown in Figure \ref{fig:tradeoff_recency_tiny} corroborates this conclusion by showing that both the working model and the semantic memory of BiRT have better stability and trade-off values compared to the baseline.

Given that TinyImageNet is one of the challenging benchmarks used in our study, we can see a very high task recency bias in DyTox in Figure \ref{fig:tradeoff_recency_tiny}, suggesting that the model is more likely to predict classes from the last few tasks for samples during inference. The skew toward recent tasks is more pronounced in the TinyImageNet data set compared to CIFAR-100. On the other hand, we can see a more balanced distribution of prediction probabilities in the working model and semantic memory of BiRT. 

\section{Hyperparameters for the Empirical Results}

We provide the hyperparameters that we used in our proposed approach for different datasets and tasks in Table \ref{tab:hparam}. Two main hyperparameters in our approach are the decay parameter $\gamma$ that is used to gradually assimilate knowledge into the semantic memory of the working model with frequency $\alpha_e$ in Eq. \ref{eqn_cls} and the weighting parameters $\beta_1$ and $\beta_2$ in Eq. \ref{eqn_cr} used to enforce consistency between the working model and the knowledge consolidated in the semantic memory with respect to images from the current task and representations from the buffer memory.

\begin{table}[t]
\centering
\caption{Hyperparameters used in BiRT for different datasets and tasks.}
\label{tab:hparam}
\vskip 0.05in
\begin{small}
\begin{sc}
\begin{tabular}{@{}lcccccc@{}}
\toprule
Dataset & \# of Tasks & Buffer Size & $\gamma$ & $\alpha_e$ & $\beta_1$ & $\beta_2$\\
\midrule
\multirow{7}{*}{CIFAR-100} 
 & \multirow{2}{*}{5} & 200 & 0.0005 & 0.001 & 0.05 & 0.01 \\
 & & 500 & 0.005 & 0.003 & 0.05 & 0.01 \\ \cmidrule{2-7}
 & \multirow{4}{*}{10} & 200 & 0.001 & 0.003 & 0.05 & 0.001 \\
 &  & 500 & 0.001 & 0.003 & 0.05 & 0.001 \\
 &  & 1000 & 0.0005 & 0.0008 & 0.05 & 0.01 \\
 &  & 2000 & 0.0002 & 0.0015 & 0.05 & 0.01 \\ \cmidrule{2-7}
 & \multirow{2}{*}{20} & 200 & 0.005 & 0.001 & 0.05 & 0.08 \\
 &  & 500 & 0.0005 & 0.003 & 0.05 & 0.1 \\
\midrule
\multirow{3}{*}{TinyImageNet} & \multirow{3}{*}{10} & 500 & 0.001 & 0.003 & 0.05 & 0.01 \\ 
 & & 1000 & 0.01 & 0.0008 & 0.01 & 0.001 \\ 
 & & 2000 & 0.0001 & 0.008 & 0.01 & 0.0008 \\ 
\midrule
\multirow{3}{*}{ImageNet-100} & \multirow{3}{*}{10} & 500 & 0.0001 & 0.003 & 0.05 & 0.001 \\
 & & 1000 & 0.0001 & 0.003 & 0.05 & 0.001 \\
 & & 2000 & 0.01 & 0.005 & 0.01 & 0.001 \\
\bottomrule
\end{tabular}
\end{sc}
\end{small}
\vskip -0.1in
\end{table}

\begin{table}[!t]
\centering
\caption{Robustness of BiRT under individual noise in CIFAR-100 dataset.}
\label{tab:robust_noise}
\vskip 0.05in
\begin{small}
\begin{sc}
\begin{tabular}{lcccc}
\toprule
\textbf{} & \multicolumn{1}{c}{\textbf{BiRT w/o Noise}} & \multicolumn{1}{c}{\textbf{Supervision Noise}} & \multicolumn{1}{c}{\textbf{Representation Noise}} & \multicolumn{1}{c}{\textbf{Attention Noise}} \\
\midrule
Last Acc  & 45.89 & 49.64 & 46.43 & 49.06 \\
Adv Acc ($\epsilon$=4) & 36.52 & 37.95 & 36.64 & 38.39 \\
Adv Acc ($\epsilon$=8) & 26.17 & 26.26 & 25.64 & 27.04 \\
Nat Cor Acc & 21.82 & 24.33 & 21.07 & 24.42 \\
\bottomrule
\end{tabular}
\end{sc}
\end{small}
\vskip -0.1in
\end{table}

\section{Robustness Analysis with Individual Noise}
In order to elucidate the improvements in robustness of BiRT brought about by different noises, we conducted more experiments to ablate the same. As shown in Table \ref{tab:robust_noise}, overall, every noise proposed in this paper contributes to improving the generalization of stored representations, enabling effective CL in vision transformers. Every noise makes the model less susceptible to adversarial attacks and more robust to natural corruption on the data.

\begin{table}[t]
\centering
\caption{Comparison of performance across different noise strengths on CIFAR-100 dataset with buffer size 500.}
\label{tab:noise_strength}
\vskip 0.05in
\begin{small}
\begin{sc}
\begin{tabular}{cccccc}
\toprule
\multicolumn{2}{c}{\textbf{Supervision Noise}} & \multicolumn{2}{c}{\textbf{Representation Noise}} & \multicolumn{2}{c}{\textbf{Attention Noise}} \\
Strength (p) & Last Acc & Strength (p) & Last Acc & Strength (p) & Last Acc \\
\midrule
0.2 & 50.90 & 0.2 & 51.45 & 0.2 & 49.93 \\
0.7 & 49.85 & 0.7 & 49.27 & 0.8 & 49.82 \\
\bottomrule
\end{tabular}
\end{sc}
\end{small}
\vskip -0.1in
\end{table}

\begin{table}[!t]
\centering
\caption{Comparison of training time taken to learn one task in CIFAR-100 dataset with buffer size 500.}
\label{tab:time}
\vskip 0.05in
\begin{small}
\begin{sc}
\begin{tabular}{llll}
\toprule
\textbf{} & \multicolumn{1}{c}{\textbf{CIFAR-100}} & \multicolumn{1}{c}{\textbf{TinyImangeNet}} & \multicolumn{1}{c}{\textbf{ImageNet-100}} \\
\midrule
DyTox & $\sim$44 mins & $\sim$2 hours 10 mins & $\sim$11 hours 22 mins \\
BiRT & $\sim$45 mins & $\sim$2 hours 6 mins & $\sim$10 hours 52 mins \\
\bottomrule
\end{tabular}
\end{sc}
\end{small}
\vskip -0.1in
\end{table}

\begin{table}[!t]
\centering
\caption{Comparison between the working model and the semantic memory of BiRT for different datasets and buffer sizes.}
\label{tab:working_semantic}
\vskip 0.05in
\begin{small}
\begin{sc}
\begin{tabular}{@{}llcc@{}}
\toprule
Dataset & \begin{tabular}[|c]{@{}c@{}}Buffer\\ Size\end{tabular} & \begin{tabular}[|c]{@{}c@{}}Working\\ Model\end{tabular} & \begin{tabular}[|c]{@{}c@{}}Semantic\\ Memory\end{tabular} \\
\midrule
\multirow{2}{*}{CIFAR-100} & 500 & 50.20 \tiny{$\pm$0.67} & 50.11 \tiny{$\pm$0.75} \\
 & 1000 & 51.20 \tiny{$\pm$1.46} & 51.17 \tiny{$\pm$1.41} \\
\multirow{2}{*}{TinyImageNet} & 500 & 32.60 \tiny{$\pm$0.18} & 32.58 \tiny{$\pm$0.24} \\
 & 1000 & 38.42 \tiny{$\pm$0.34} & 38.24 \tiny{$\pm$0.37} \\
\multirow{2}{*}{ImageNet-100} & 500 & 51.06 \tiny{$\pm$0.24} & 50.80 \tiny{$\pm$0.56} \\
 & 1000 & 52.21 \tiny{$\pm$0.00} & 51.69\tiny{$\pm$0.00} \\
 \bottomrule
\end{tabular}
\end{sc}
\end{small}
\vskip -0.1in
\end{table}

\section{Sensitivity Analysis to Noise}
We control the strength and amount of noise added at different stages of the training process, based on the percentage of samples to which noise is added in each batch. We conducted additional experiments on CIFAR-100 with 10 tasks and a buffer size of 500, varying the percentage of samples to which each noise type is added. The results are shown in Table \ref{tab:noise_strength}. `p' denotes the percentage of samples to which the corresponding noise is added during the replay of the representation in each batch (batch\_size = 128). It is evident that different levels of noise change the last accuracy; however, the performance at different levels of noise reveals that BiRT is not very sensitive to hyperparameters.

\section{Training Time Analysis}
We conducted an experiment to compare the training time of different CL models considered in this work. The training time on an NVIDIA RTX 2080 Ti for various datasets with buffer size 500 to learn a single task (500 epochs) in CIFAR-100 dataset is enumerated in Table \ref{tab:time}. As can be seen, both DyTox and BiRT entail similar training times, indicating that the proposed noise-based approach in BiRT does not increase the training time. In fact, our proposed approach improves generalization performance to a large extent with minimal/no additional computational cost.

\section{Analysis on Working Model and Semantic Memory}
We compare the performance between DyTox and the BiRT working model in Table \ref{tab:main}. However, stochastically assimilating the knowledge learned in the working model into the semantic memory throughout the learning process and at the end of tasks results in a generalized working model with lesser forgetting. We show the last accuracy of the working model and semantic memory for different datasets and buffer sizes in Table \ref{tab:working_semantic}.

\begin{table}[t]
\centering
\caption{Quantitative results for the stability-plasticity analysis of different CL models.}
\label{tab:plasticity_num}
\vskip 0.05in
\begin{small}
\begin{sc}
\begin{tabular}{lccc}
\toprule
\textbf{} & \textbf{Plasticity} & \textbf{Stability} & \textbf{Trade-off} \\
\midrule
DyTox & 73.30 & 18.08 & 29.01 \\
BiRT & \textbf{43.73} & \textbf{32.75} & \textbf{37.45} \\
\bottomrule
\end{tabular}
\end{sc}
\end{small}
\vskip -0.1in
\end{table}

\begin{table}[t]
\centering
\caption{Quantitative results of different CL models to different levels of adversarial attacks. Noise in BiRT improves its robustness against adversarial attacks across different epsilon values.}
\label{tab:adversarial_num}
\vskip 0.05in
\begin{small}
\begin{sc}
\begin{tabular}{lcccccccccc}
\toprule
\textbf{} & \textbf{Average} & \multicolumn{1}{c}{\textbf{0}} & \multicolumn{1}{c}{\textbf{0.25}} & \multicolumn{1}{c}{\textbf{0.5}} & \multicolumn{1}{c}{\textbf{1}} & \textbf{2} & \textbf{4} & \textbf{8} & \textbf{16} & \textbf{32} \\
\midrule
DyTox & 23.59 & 35.29 & 33.35 & 31.43 & 29.30 & 28.26 & 27.88 & 19.16 & 7.03 & 0.67 \\
BiRT w/o Noise & 31.50 & 47.45 & 44.74 & 41.77 & 38.92 & 37.54 & 36.53 & 26.18 & 9.54 & 0.90 \\
BiRT & \textbf{33.37} & \textbf{50.27} & \textbf{47.32} & \textbf{44.23} & \textbf{41.30} & \textbf{39.79} & \textbf{38.95} & \textbf{27.37} & \textbf{10.12} & \textbf{1.03} \\
\bottomrule
\end{tabular}
\end{sc}
\end{small}
\vskip -0.1in
\end{table}

\begin{table}[!t]
\centering
\caption{Quantitative results of different CL models to different levels of natural corruption. Noise in BiRT improves its robustness against natural corruption across different strengths.}
\label{tab:natural_num}
\vskip 0.05in
\begin{small}
\begin{sc}
\resizebox{\textwidth}{!}{
\begin{tabular}{lcccccccccc}
\toprule
\textbf{} & \textbf{Average} & \textbf{bright.} & \textbf{contrast} & \textbf{defocus} & \textbf{elastic} & \textbf{fog} & \textbf{frost} & \textbf{g\_blur} & \textbf{g\_noise} & \textbf{glass} \\
\cmidrule{2-11}
DyTox & 21.06 & 26.66 & 12.38 & 22.64 & 21.32 & 18.57 & 24.92 & 21.21 & 20.99 & 17.08 \\
BiRT w/o Noise & 21.82 & 28.71 & 10.99 & 22.19 & 21.34 & 16.07 & 27.44 & 20.40 & 23.41 & 18.32 \\
BiRT & \textbf{25.81} & \textbf{32.59} & \textbf{14.19} & \textbf{26.49} & \textbf{25.53} & \textbf{20.04} & \textbf{32.46} & \textbf{24.34} & \textbf{27.17} & \textbf{22.22} \\
\cmidrule{2-11}
 & \textbf{impulse} & \textbf{jpeg} & \textbf{motion} & \textbf{pixelate} & \textbf{saturate} & \textbf{shot} & \textbf{snow} & \textbf{spatter} & \textbf{speckle} & \textbf{zoom} \\
\cmidrule{2-11}
DyTox & 18.51 & 22.69 & 19.56 & 24.54 & 21.60 & 22.00 & 21.39 & 22.43 & 21.11 & 20.57 \\
BiRT w/o Noise & 19.83 & 25.23 & 18.22 & 24.25 & 21.36 & 24.36 & 25.05 & 24.57 & 23.30 & 19.71 \\
BiRT & \textbf{22.86} & \textbf{29.71} & \textbf{21.84} & \textbf{28.71} & \textbf{24.68} & \textbf{28.32} & \textbf{29.58} & \textbf{28.67} & \textbf{27.09} & \textbf{23.98} \\
\bottomrule
\end{tabular}}
\end{sc}
\end{small}
\vskip -0.1in
\end{table}

\section{Quantitative Results for Model Analysis}
Figure \ref{fig:tradeoff_recency} in the main text illustrates the stability-plasticity trade-off between DyTox and BiRT. We provide the quantitative results for the same in Table \ref{tab:plasticity_num}. DyTox is more prone to forgetting, whereas BiRT displays a better stability-plasticity trade-off compared to
the baseline. We evaluated the robustness of DyTox, BiRT without noise, and BiRT across different strengths of adversarial attacks and natural corruptions. Qualitative results are presented in Figure \ref{fig:robustness} in the main text. Table \ref{tab:adversarial_num} and \ref{tab:natural_num} enumerate the quantitative results of the same.

\section{Extended Related Works}
In addition to the CL methods discussed in the Related Works section in the main text, there is another line of work that pursues the `Deep Inversion' technique to synthesize replay images for old tasks. Deep inversion works by inverting a neural network's feature extractor to generate synthetic input data that is similar to the original input data. In the context of class incremental learning, deep inversion can be used to generate synthetic data for the new classes that the model needs to learn without requiring access to any real data for those classes \citep{yin2020dreaming, gao2022r, smith2021always}. Though this approach alleviates any privacy issues and is more memory-efficient, the model responsible for generating the synthetic data might undergo catastrophic forgetting and this can be exacerbated in long-task sequences.

The theory of a complementary learning system (CLS) posits that the ability to continually acquire and assimilate knowledge over time in the brain is mediated by multiple memory systems \citep{hassabis2017neuroscience, kumaran2016learning}. Inspired by CLS theory, CLS-ER \citep{arani2022learning} proposed a dual memory method that maintains multiple semantic memories that interact with episodic memory. On the other hand, FearNet \citep{kemker2017fearnet} utilizes a brain-inspired dual-memory system coupled with pseudo rehearsal \citep{robins1995catastrophic} in order to efficiently learn new tasks.

\section{Limitations}
BiRT is a novel continual learning approach that can be applied to various tasks. However, the effectiveness of different levels of noise in BiRT varies in terms of generalization and robustness. The impact of hyperparameters on the effectiveness of different types of noise can also affect accuracy to some extent. However, our empirical results reveal that BiRT is not very sensitive to hyperparameters. BiRT may not be well-suited for datasets with small images (e.g., 32 x 32) since the representations stored in the buffer for such datasets may require more memory compared to storing images. Nonetheless, since real-world datasets typically contain high-resolution images (as in ImageNet-100 and TinyImageNet), BiRT can enable efficient CL in most cases. BiRT does not raise privacy concerns as we do not store personal data, and there are no known bias and fairness issues since we do not use any pretrained weights.

\section{Attention Map Analysis}\label{attention_maps}
A CL model that is able to preserve the salient regions learned in the first task (when those samples were trained) as learning progresses through the subsequent tasks would provide less catastrophic forgetting \citep{ebrahimi2021remembering}. 
The [CLS] token in Vision Transformers, which is utilized to infer the class of a sample \citep{dosovitskiy2020image}, attends to the salient regions of an image in order to extract rich features pertaining to the task learned by the model. Therefore, it would be beneficial to study the drift in the regions that the model considers to be salient in the image as learning progresses. 

\begin{figure*}[t]
\begin{center}
\vskip 0.2in
\centerline{
\begin{tabular}{c}
    \includegraphics[width=.75\textwidth]{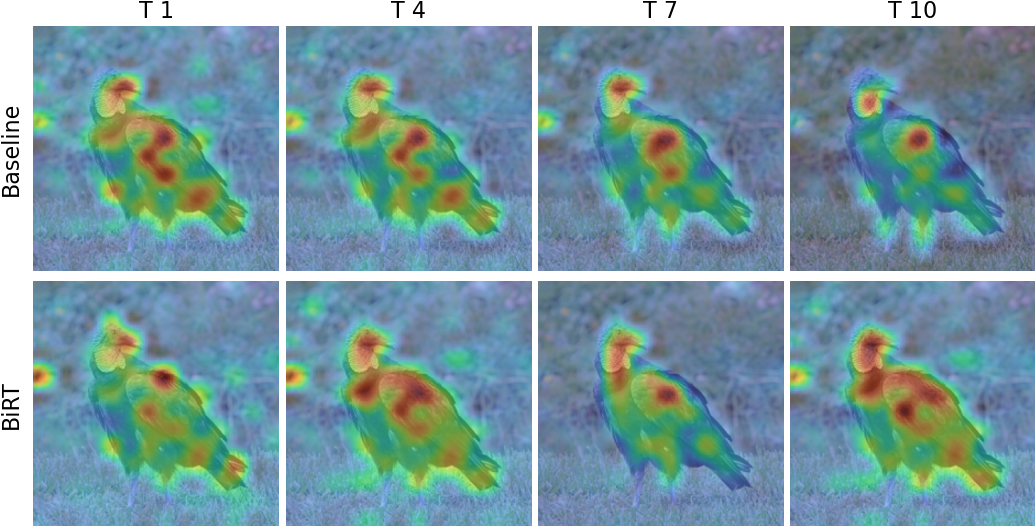} \\
    \includegraphics[width=.75\textwidth]{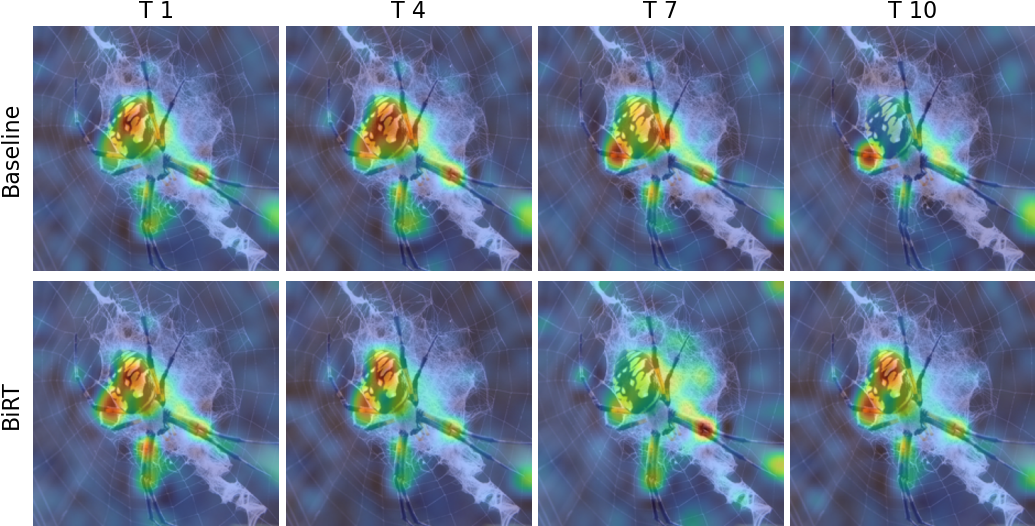} \\
\end{tabular}}
\caption{Comparison of attention maps with respect to the class token on the validation set of the first task of ImageNet-100 trained for 10 tasks with buffer size 500. The attention maps are plotted after learning the first, fourth, seventh, and last tasks (red regions correspond to regions with higher attention). BiRT retains the knowledge of salient regions in the image better than DyTox, leading to better predictions and less forgetting.}
\label{fig:attention2}
\end{center}
\vskip -0.2in
\end{figure*}

\begin{figure*}[b]
\begin{center}
\vskip 0.2in
\centerline{
\begin{tabular}{c}
    \includegraphics[width=.65\textwidth]{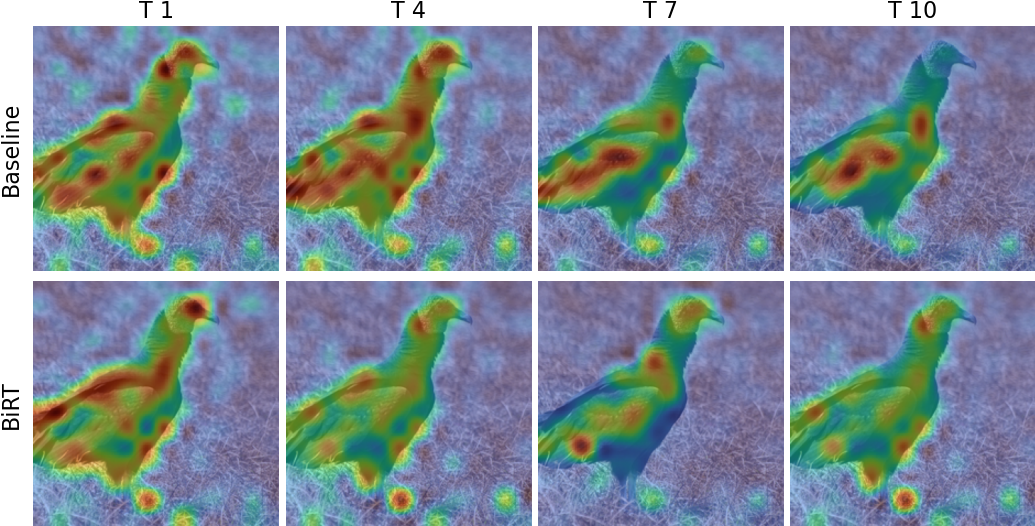} \\
    \includegraphics[width=.65\textwidth]{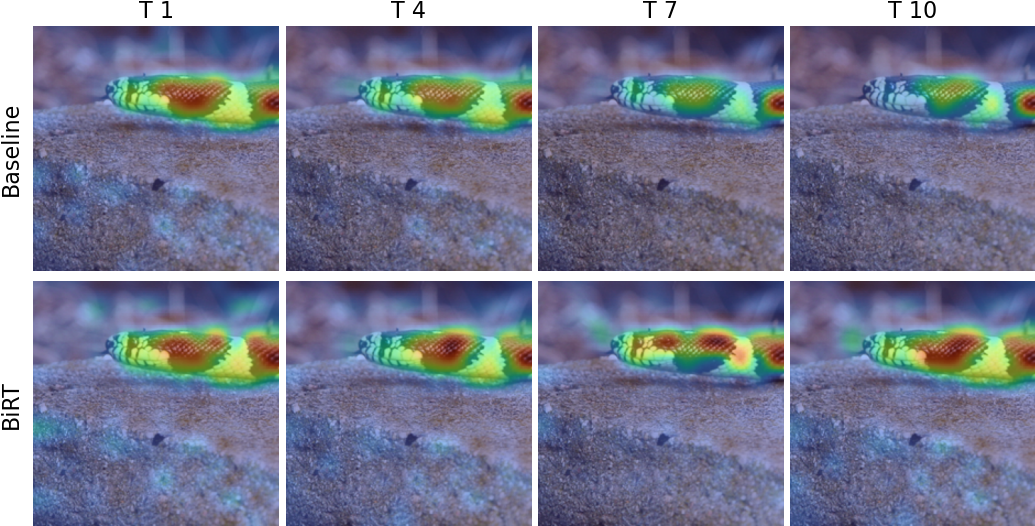} \\ 
    \includegraphics[width=.65\textwidth]{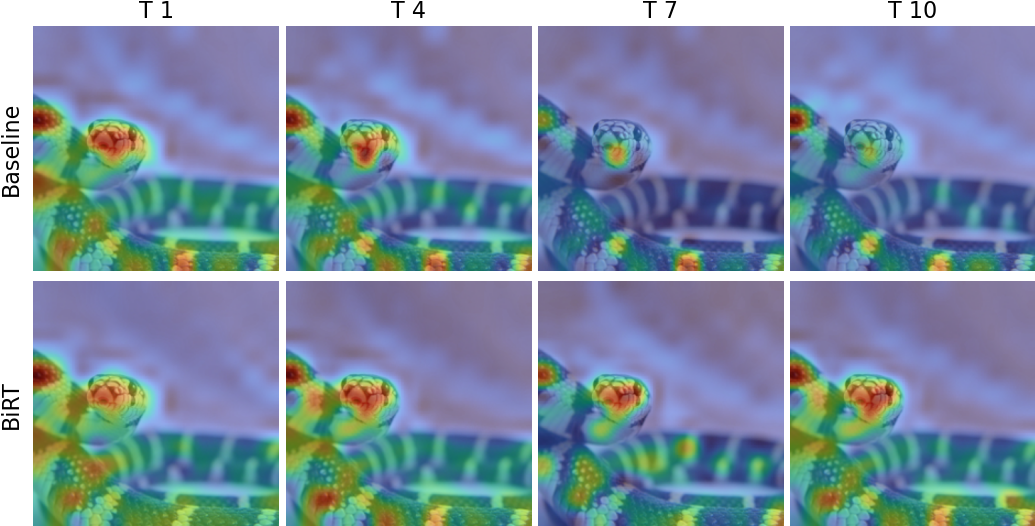} \\
\end{tabular}}    
\caption{Comparison of attention maps with respect to the class token on the validation set of the first task of ImageNet-100 trained for 10 tasks with buffer size 500. The attention maps are plotted after learning the first, fourth, seventh, and last tasks (red regions correspond to regions with higher attention). BiRT retains the knowledge of salient regions in the image better than DyTox, leading to better predictions and less forgetting.}
\label{fig:attention3}
\end{center}
\vskip -0.2in
\end{figure*}

Concretely, we study the attention maps calculated by the last Class-Attention block in BiRT for samples in the validation set of the first task as the learning progresses from the first task to the last task. We overlay the attention map as a heatmap (interpolated to the image size) on the image. Figures \ref{fig:attention2} and \ref{fig:attention3} show that the BiRT working model preserves the attention map learned in the first task better than DyTox as the training progresses.

\end{document}